%% file: neurips_2026.tex
\documentclass{article}

\usepackage[preprint]{neurips_2026}

\usepackage[utf8]{inputenc} 
\usepackage[T1]{fontenc}    
\usepackage{hyperref}       
\usepackage{url}            
\usepackage{booktabs}       
\usepackage{amsfonts}       
\usepackage{nicefrac}       
\usepackage{microtype}      
\usepackage{xcolor}         

\usepackage{wrapfig}
\usepackage{graphicx}
\usepackage{fontawesome} 
\usepackage{makecell}
\usepackage{wrapfig}  
\usepackage{adjustbox}
\usepackage{mathrsfs}
\usepackage{listings} 
\usepackage{pifont}
\usepackage{subfigure}
\usepackage{alltt}  
\usepackage{subcaption}
\usepackage{svg}
\usepackage{makecell}
\usepackage{amsmath}
\usepackage{amsfonts}
\usepackage{amssymb}
\usepackage{cleveref}
\crefname{section}{§}{§§}
\Crefname{section}{§}{§§}
\usepackage{svg}
\usepackage[normalem]{ulem} 
\usepackage{colortbl}
\usepackage{multirow}
\usepackage{algorithm}
\usepackage{algpseudocode}
\usepackage[figuresright]{rotating}
\usepackage[most]{tcolorbox}
\usepackage{paralist}
\usepackage{enumitem}  

\newcommand{\cmark}{\textcolor{green!60!black}{\checkmark}}
\newcommand{\xmark}{\textcolor{red}{\ding{55}}}
\newcommand{\tmark}{\textcolor{blue}{\faToggleOn}} 

\definecolor{BoxBackground}{RGB}{240, 240, 240} 
\definecolor{BoxFrame}{RGB}{0, 0, 0} 
\definecolor{TitleBackground}{RGB}{0, 0, 0} 
\definecolor{TitleText}{RGB}{255, 255, 255} 
\tcbset{
  academicbox/.style={
    boxsep=5pt,
    left=2pt,
    right=2pt,
    bottom=0.5pt,
    boxrule=0.5pt,
    colback=BoxBackground,
    colframe=BoxFrame,
    colbacktitle=TitleBackground,
    coltitle=TitleText,
    enhanced,
    attach boxed title to top left={yshift=-0.1in,xshift=0.1in},
    boxed title style={boxrule=0pt,colframe=white},
    title={#1},
  }
}
\newtcolorbox{AcademicBox}[1][]{academicbox=#1}

\definecolor{s_doc_qa_c}{HTML}{da0d68}
\definecolor{m_doc_qa_c}{HTML}{da1d23}
\definecolor{summarization_C}{HTML}{ebb40f}
\definecolor{dialogue_c}{HTML}{187a2e}
\definecolor{synthetic_c}{HTML}{0aa3b5}

\title{\textsc{MMLongCite}: A Benchmark for Evaluating Faithfulness of Long-Context Vision-Language Models}

\author{%
    Keyan Zhou${}^{1,2}$\thanks{Work done during an internship at ByteDance.} \qquad Zecheng Tang${}^{1}$ \qquad Lingfeng Ming${}^{2}$ \qquad Qiguang Chen${}^{3}$ \\
    \textbf{Wangjie You${}^{1}$ \qquad Guanghao Zhou${}^{4}$ \qquad Dan Qiao${}^{2}$ \qquad Zheming Yang${}^{2}$} \\
    \textbf{Libo Qin${}^{5}$ \qquad Minghui Qiu${}^{2}$ \qquad Juntao Li${}^{1}$ \qquad Min Zhang${}^{1}$} \\
    ${}^1$Soochow University \quad ${}^2$ByteDance \\ \quad ${}^3$Harbin Institute of Technology \quad ${}^4$East China Normal University \\ ${}^5$Harbin Institute of Technology, Shenzhen \\
    \{kyzhou49, zctang\}@stu.suda.edu.cn \qquad \{ljt, minzhang\}@suda.edu.cn \\ qinlibo@hit.edu.cn
}

\begin{document}

\maketitle

\begin{abstract}

The rapid advancement of long-context vision language models (LCVLMs) has led to a significant expansion of their context windows. However, an extended context window does not guarantee the effective utilization of the context, posing a critical challenge for real-world applications. Current evaluations of such long-context faithfulness in multimodal settings remain limited to short contexts. To bridge this gap, we introduce \textsc{MMLongCite}, the first benchmark evaluating the faithfulness of LCVLMs via multimodal citation generation.
\textsc{MMLongCite} features 2,280 examples across 8 tasks and diverse modalities (image, video, interleaved), with context lengths scaled from 16K to 128K tokens.
To test spatial localization capabilities of LCVLMs, we also introduce \textsc{MMLongCite-HR}, evaluating fine-grained visual grounding amidst dense pixel spaces.
Through extensive benchmarking of cutting-edge LCVLMs, we provide a systematic analysis of current multimodal citation capabilities.
Our results reveal a significant discrepancy between answer correctness and citation faithfulness.
We also conduct attention pattern investigations and in-depth error analyses to reveal the underlying phenomena of failures in LCVLMs.
\textsc{MMLongCite}\setcounter{footnote}{0}\footnote{The code and data are available at \url{https://github.com/bytedance/MMLongCite}.} establishes a rigorous foundation for diagnosing and advancing the faithfulness of LCVLMs. We hope our findings provide meaningful insights to drive further improvements in the long-context capabilities of LCVLMs.

\end{abstract}

\input{section/introduction}

\input{section/related_work}

\input{section/method}

\input{section/experiment}

\input{section/analysis}
\input{section/conclusion}

\section*{Acknowledgements}
We thank the anonymous reviewers for their valuable comments.
Libo Qin, Minghui Qiu, and Juntao Li are the corresponding authors.
This work was supported by the National Natural Science Foundation
of China (NSFC) under grants 92570120 and 62306342, and by the
collaborative project ``Research and Exploration of E-commerce
Auditing Agents Based on Multimodal Large Models,'' jointly
conducted by Soochow University and ByteDance. The first author thanks Lingfeng Ming for his mentorship and helpful discussions on this work during the internship at ByteDance.

\bibliographystyle{plain}
\bibliography{custom}
\input{section/appendix}


\end{document}

%% file: section/introduction.tex
\begin{figure*}[t]
    \centering
    \includegraphics[width=\linewidth]{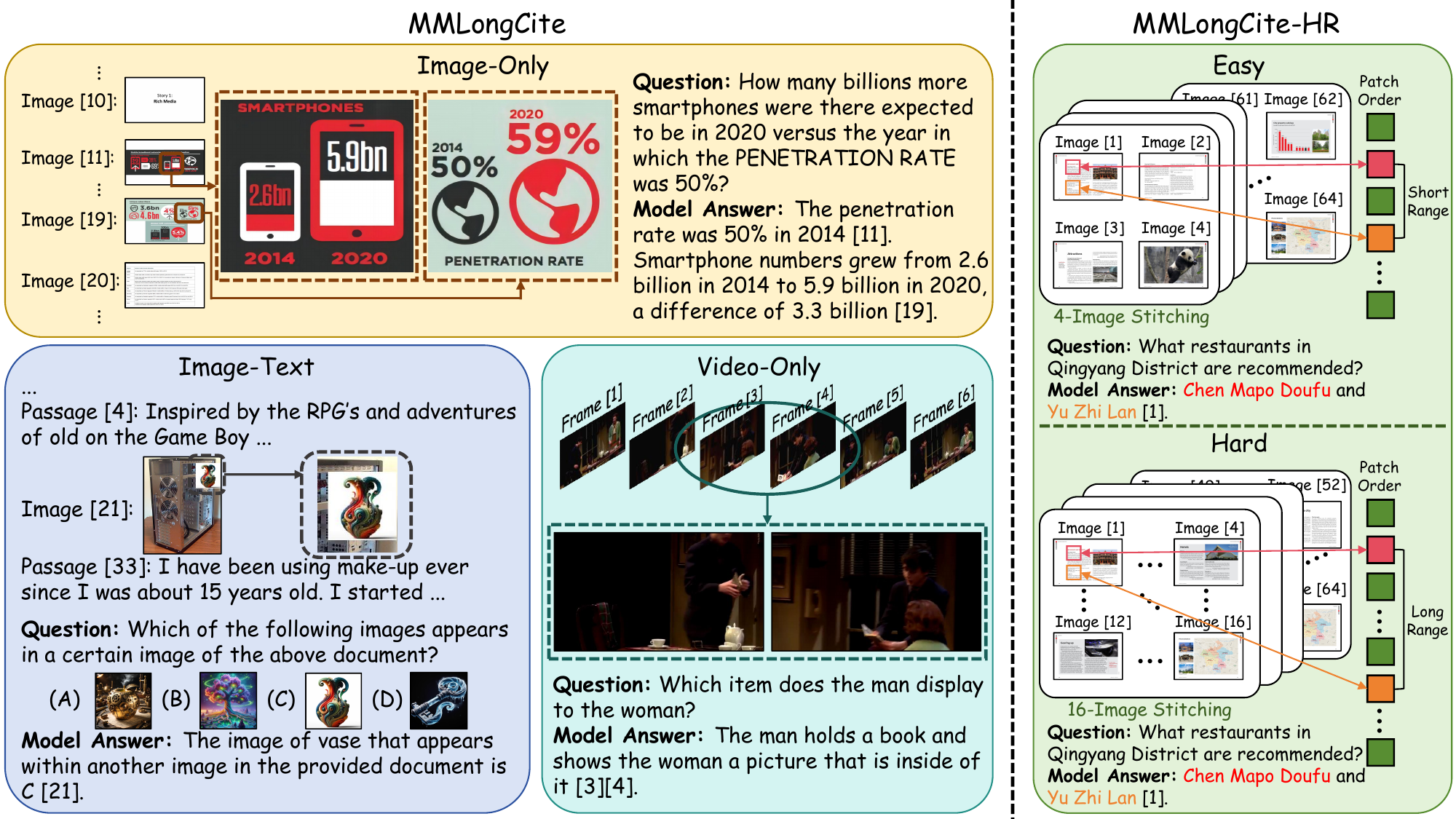}
    \caption{An overview of our proposed \textbf{\textsc{MMLongCite}} benchmark.
(Left) The primary task evaluates the processing of multimodal long contexts in Image-Only, Image-Text, and Video-Only formats.
(Right) \textbf{\textsc{MMLongCite-HR}} assesses visual grounding capabilities with scalable difficulty. In the Easy setting, models are fed a sequence of composite images, where each composite is stitched from 4 source images. In the Hard setting, models are fed a sequence of composite images stitched from 16 source images.}
    \label{fig:intro_diff}
\end{figure*}

\section{Introduction}

Recent advancements in long-context vision-language models (LCVLMs) have significantly extended their context windows, enabling them to process vast, interleaved streams of text, images and videos~\citep{lu2024text,comanici2025gemini,ge2024v2pe}. However, a large window size does not guarantee faithful utilization of the long context. A critical challenge persists: when there is a conflict between the provided context and the model's internal parametric knowledge, the model often prioritizes its own knowledge, disregarding the context and leading to hallucinated responses~\citep{tang2025citeeval}. This lack of faithfulness undermines LCVLM reliability, severely limiting their practical deployment in scenarios requiring alignment with time-sensitive or dynamic contextual information.

To enable the quantification of the  faithfulness for cutting-edge LCVLMs, a prominent approach is to require models to generate citations alongside their responses~\citep{gao2023enabling,hu2025mcitebench,zhang2025longcite, tang2025citeeval,wu2025ref}. This not only compels the model to ground its output in the provided context but also empowers users with a verifiable pathway to trace the information to its source, thus facilitating a direct assessment of the response's reliability.
Despite recent progress in developing benchmarks for multimodal faithfulness~\citep{hu2025mcitebench}, existing efforts still exhibit substantial limitations, including (1) \textbf{Insufficient Context Length}, failing to genuinely challenge the long-context capabilities of LCVLMs; (2) \textbf{Lack of Context Diversity}, as they are confined to a single document type, i.e., academic papers; (3) \textbf{Limited Task 
Categories}, with task formats being confined to explanation and localization, thereby failing to assess more complex cross-modal reasoning abilities.

To bridge these gaps and rigorously evaluate the faithfulness of LCVLMs, we introduce \textbf{\textsc{MMLongCite}}, the first benchmark specifically designed for long-context multimodal citation, unifying standard visual QA with the prevalent vision-text compression QA~\citep{zhao2025vtcbench,wei2025deepseek,cheng2025glyph}. 
\textsc{MMLongCite} is characterized by three key properties that address the aforementioned limitations: (1) Context Scalability: To genuinely challenge LCVLMs, context lengths are scaled from 16K to 128K tokens across 4 intervals. (2) Modality Diversity: It incorporates a rich variety of data sources—including documents, natural images, videos and supports three complex context formats (image-only, interleaved image-text, and video-only), ensuring coverage across varied information sources, as shown in Figure~\ref{fig:intro_diff}.
(3) Task Complexity: It features a diverse set of 4 broad categories and 8 distinct tasks, varied task levels ranging from single-hop extraction~\citep{deng-etal-2025-longdocurl,wuvisual} to complex multi-hop QA~\citep{tanaka2023slidevqa,ho2020constructing}.
To further probe the limits of fine-grained visual grounding, we introduce \textbf{\textsc{MMLongCite-HR}} (High-Resolution). By aggregating independent images into a unified, high-resolution grid, we shift the challenge from sequential retrieval to spatial precision. This configuration demands a granular focus mechanism capable of isolating specific visual evidence within a dense pixel space without being distracted by adjacent visual contexts. This effectively simulates real-world scenarios requiring synthesis across simultaneous visual inputs, such as radiologists comparing concurrent scans or security personnel monitoring a surveillance matrix.

In summary, our contributions are as follows:
\begin{itemize}[leftmargin=*, topsep=0pt, itemsep=2pt]
    \item We introduce \textsc{MMLongCite}, the first benchmark specifically designed to evaluate the multimodal citation faithfulness of LCVLMs across diverse long contexts.
    
    \item We benchmark 28 cutting-edge LCVLMs, uncovering a correctness-faithfulness gap and demonstrating that small models are prone to generating hallucinated citations rather than grounding responses in the provided context. We also assess the impact of spatially dense, high-resolution inputs on the faithfulness of models.
    
    \item Through an in-depth analysis of attention patterns during citation generation, we dissect the fundamental failure modes underlying unfaithful responses, offering a rigorous diagnostic foundation for future LCVLM development.
\end{itemize}

%% file: section/related_work.tex
\section{Related Work}

\subsection{Faithfulness and Citation Generation}

Ensuring faithfulness and mitigating hallucination remains a central challenge for large language and vision-language models, with attribution emerging as a crucial mitigation strategy \cite{li2023survey}. Pioneering works like ALCE \cite{gao2023enabling} laid the foundation, followed by evaluations specifically tailored for long contexts, such as LongCite \cite{zhang2025longcite}, L-CiteEval \cite{tang2025citeeval}, Ref-Long \cite{wu2025ref}, and query-focused summarization datasets \cite{wright-etal-2025-unstructured, deng2024webcites}. Driven by these benchmarks, extensive methods have been developed to improve citation generation. These include self-supervised alignment \cite{chuang2025selfcite, wang2025A3}, preference learning \cite{li2024improving}, and fine-grained rewards \cite{huang2024training}. Furthermore, researchers have systematically analyzed citation failures \cite{buchmann2025citation} and explored generation-time versus post-hoc correction strategies \cite{maheshwari2025citefix, saxena2025generation}. 

However, these evaluations and methodologies are predominantly confined to textual sources, leaving multimodal attribution nascent. While MCiteBench \cite{hu2025mcitebench} represents a pioneering step, its evaluation is constrained by a singular data domain (academic papers), a limited set of simple tasks, and context lengths that do not sufficiently tax state-of-the-art LCVLMs. \textsc{MMLongCite} addresses these shortcomings by introducing diverse data modalities, complex cross-modal tasks, and significantly longer contexts to establish a more comprehensive and rigorous standard for multimodal faithfulness.

\subsection{Long-Context Multimodal Benchmark}

Existing multimodal long-context benchmarks evaluate LCVLMs primarily through distinct task domains. In the document domain, benchmarks like LongDocURL~\cite{deng-etal-2025-longdocurl}, MMLongBench-Doc~\cite{ma2024mmlongbench}, and M-Longdoc~\cite{chia2025m} focus on long document understanding, challenging models to perform cross-page reasoning and fine-grained information extraction from visual inputs, while MMDocIR~\cite{dong2025mmdocir} further extends this line to page-level retrieval over lengthy documents. Diagnostic tasks such as MM-NIAH~\cite{wang2024needle}, MMNeedle~\cite{wang2025multimodal}, and Visual Haystack~\cite{wuvisual} assess precise information retrieval via multimodal ``Needle-In-A-Haystack'' tests, with follow-up efforts~\cite{sharma2024losing,zhao2025vtcbench,zhaoneedle,ye2025re} extending this paradigm to visually compressed text and temporal search over long videos. Long-form video understanding is further targeted by benchmarks such as Video-MME~\cite{fu2025video}, LongVideoBench~\cite{wu2024longvideobench}, and MLVU~\cite{zhou2025mlvu}, emphasizing temporal modeling and event-level reasoning over hour-scale content. Beyond vision, recent benchmarks~\cite{he2025audiomarathon,luo2026chronosaudio} extend long-context evaluation to the audio modality, probing temporal reasoning over extended speech and acoustic scenes. While comprehensive suites~\cite{wangmmlongbench,dingjie2024milebench} aggregate diverse multimodal long-context tasks, they focus solely on answer correctness, overlooking whether the rationale is grounded in the provided context.

%% file: section/method.tex
\section{MMLongCite}
Our benchmark, \textsc{MMLongCite}, comprises 4 distinct tasks: Single-Source Visual QA, Multi-Source Visual QA, Visual Grounding and Video Understanding. It integrates 8 diverse datasets and encompasses 2,280 samples, with the length of the context ranging from 16K to 128K. We use the Qwen2.5-VL processor to calculate the number of text and visual tokens in the context. The statistics of dataset can be viewed in Table~\ref{tab:data_statistic}.

\subsection{Problem Definition}
Given the multimodal long context $D$ and a user query $Q$, the model is required to generate the response $R$, which is composed of $n$ statements $\mathcal{S}=\{s_1, s_2, \cdots, s_{n}\}$ and their corresponding citations $\mathcal{C}_i=\{c_{i,1}, c_{i,2}, \cdots\}$. To enable the model to accurately generate citations, we assign a unique citation index to each individual image, frame, or text passage within the context. For example, the model needs to generate a response like "Water boils at 100°C at sea level [3][5].", where the specific citation indices must be explicitly displayed in box brackets.

\input{table/data_statistics}

\subsection{Task Setup and Benchmark Construction}
We develop \textsc{MMLongCite} by rigorously filtering and repurposing 8 datasets that align with our task objectives, adopting a postprocessing strategy similar to \citep{wangmmlongbench}. The benchmark is organized into three modality-based tracks: image-only, interleaved image-text, and video-only, as depicted in Figure~\ref{fig:intro_diff}. A breakdown of dataset statistics is provided in Table~\ref{tab:data_statistic} and refer to Appendix~\ref{appendix:benchmark_construction} for more details on the benchmark construction pipeline.
The limited sample size within each context length interval
of each dataset may constrain the benchmark's distributional
representativeness. We discuss this limitation in Appendix~\ref{appendix:statistical_uncertainty}
and~\ref{appendix:limitation}.

\subsubsection{Image-Only Contexts}
This category evaluates the model to process long contexts composed entirely of images. It comprises five datasets: LongDocURL~\citep{deng-etal-2025-longdocurl}, MMLongBench-Doc~\citep{ma2024mmlongbench}, SlideVQA~\citep{tanaka2023slidevqa}, 2WikiMultiHopQA~\citep{ho2020constructing}, and Visual Haystack~\citep{wuvisual}. The construction method varies depending on whether the task involves a single, continuous document or a collection of discrete images.

For tasks based on a single, cohesive document (LongDocURL and MMLongBench-Doc), the goal is to assess deep comprehension within a continuous visual narrative. These tasks encompass both single-page and multi-page evidence QA scenarios. To ensure a balanced distribution across various context lengths, we normalize each source document to a target length $L_t$. To preserve the original positional bias of the key information, which is known to influence model performance~\citep{liu2024lost}, our resizing strategy proportionally trims the prefix and suffix context surrounding the evidence region. The new lengths for the prefix ($L'_{p}$) and suffix ($L'_{s}$) are calculated as:
\begin{equation}
\begin{aligned}
    L'_{p} &= (L_{t} - L_{e}) \times \frac{L_{p}}{L_{p} + L_{s}} \\
    L'_{s} &= (L_{t} - L_{e}) \times \frac{L_{s}}{L_{p} + L_{s}},
\end{aligned}
\end{equation}
where $L_e$, $L_p$, and $L_s$ represent the lengths of the evidence, original prefix and suffix, respectively.

For tasks requiring information aggregation from discrete sources (SlideVQA and 2WikiMultiHopQA for multi-hop reasoning, and Visual Haystack for single-hop retrieval), we create long contexts by placing evidence images within a long sequence of distractor images. For the text-native multi-hop dataset 2WikiMultiHopQA, we first convert all text documents into images.
Then we employ a unified pipeline to postprocess the three datasets for constructing long-context visual inputs. Formally, let a source dataset be represented as a collection of triplets $\mathcal{D} = \{(q, a, \mathcal{E})\}$, where $q$ denotes the question, $a$ is the ground-truth answer, and $\mathcal{E} = \{e_1, e_2, \dots, e_k\}$ represents the set of essential evidence images required to derive $a$.

For each instance, we first sample a seed triplet $(q, a, \mathcal{E})$ from the source dataset. To simulate long-context scenarios with varying noise levels, we introduce a distractor set $\mathcal{R}$, consisting of irrelevant images sampled from the same domain. We then sample a subset of distractors $\mathcal{D}_{\text{dist}} \subset \mathcal{R}$ such that the combined sequence length $|\mathcal{E}| + |\mathcal{D}_{\text{dist}}|$ equals a target context length $L_{t}$. The final input sequence $S$ is constructed by combining the evidence set $\mathcal{E}$ and the distractor set $\mathcal{D}_{\text{dist}}$ together. This process ensures that the evidence is randomly distributed within the long visual context, forcing the model to retrieve and aggregate information amidst significant noise, rather than relying on positional biases.

\subsubsection{Interleaved Image-Text Contexts}
This category, represented by MM-NIAH~\citep{wang2024needle}, tests a model's ability to process long, heterogeneous contexts where images and text passages are interspersed. The core task is to retrieve a specific piece of information, which can be either an image or a text snippet, from this mixed-modality sequence. We adapt instances from the \texttt{retrieval-image} and \texttt{retrieval-text} subtasks. We sample a uniform number of instances from four distinct length intervals for evaluation.

\subsubsection{Video-Only Contexts}
This category focuses on temporal reasoning and is built upon Video-MME~\citep{fu2025video} and LongVideoBench~\citep{wu2024longvideobench}. A primary challenge with long videos is that the high number of frames (e.g., at 30 frames per second) generates sequences that vastly exceed the context capacity of current models. To address this, we standardize all videos by downsampling them to fixed frame per second (fps). From these downsampled frame sequences, we then sample instances of varying lengths to construct the final evaluation set.

\subsubsection{\textsc{MMLongCite-HR}}
To explicitly evaluate visual grounding and spatial reasoning capabilities, we introduce \textsc{MMLongCite-HR} (High-Resolution), constructed from LongDocURL, 2WikiMultiHopQA, Visual Haystack, and LongVideoBench within \textsc{MMLongCite}. This extension challenges models to operate on composite inputs and is structured into two difficulty levels, as illustrated in Figure~\ref{fig:intro_diff}.

In the Easy setting, every 4 context images are consolidated into a single medium-resolution composite (approximately 1K--2K resolution), whereas the Hard setting aggregates 16 images into a high-resolution layout (approximately 2K--4K resolution). Unlike \textsc{MMLongCite} which inputs images as discrete, sequential instances, \textsc{MMLongCite-HR} compels the model to implicitly segment high-resolution composite views to localize specific sub-regions and map evidence back to citation indices. This design facilitates a rigorous assessment of fine-grained spatial reasoning and information attribution within high-resolution long-context scenarios. \textsc{MMLongCite-HR} scores reflect a combined measure of spatial
reorganization robustness and high-resolution visual grounding, rather than
a pure measure of high-resolution visual understanding.
Further construction details and analysis are provided in
Appendix~\ref{appendix:construction_sensitivity}.

%% file: table/data_statistics.tex
\begin{table*}[t]
    \centering
    \caption{Statistics of tasks in \textsc{MMLongCite}.}
    \resizebox{1\textwidth}{!}{
    \begin{tabular}{ l l c c c c c c}
        \toprule
         \multirow{2}{*}{\bf Tasks} & \multirow{2}{*}{\bf Source} & \multirow{2}{*}{\bf \makecell[c]{Context}} &  \multicolumn{4}{c}{\bf Length Distribution} & \multirow{2}{*}{\bf Total} \\
         \cmidrule{4-7}
         & & & \bf 0$\sim$16k & \bf 16$\sim$32k & \bf 32$\sim$64k & \bf 64$\sim$128k\\
         \midrule
         \rowcolor{s_doc_qa_c!25} \multicolumn{8}{c}{\textit{\textbf{Single-Source Visual QA}}}  \\
         LongDocURL & \citep{deng-etal-2025-longdocurl} & Image & 80 & 80 & 80 & 80 & 320\\
         MMLongBench-Doc & \citep{ma2024mmlongbench} & Image & 80 & 80 & 80 & 80 & 320\\
         \midrule
         \rowcolor{m_doc_qa_c!25} \multicolumn{8}{c}{\textit{\textbf{Multi-Source Visual QA}}} \\
         SlideVQA & \citep{tanaka2023slidevqa} & Image & 70 & 70 & 70 & 70 & 280\\
         2WikiMultiHopQA & \citep{ho2020constructing} & Text$\rightarrow$Image & 80 & 80 & 80 & 80 & 320\\
         \midrule
         \rowcolor{summarization_C!25} \multicolumn{8}{c}{\textit{\textbf{Visual Grounding}}} \\
         Visual Haystack & \citep{wuvisual} & Image & 100 & 100 & 100 & 100 & 400\\
         MM-NIAH & \citep{wang2024needle} & Image, Text & 80 & 80 & 80 & 80 & 320\\
         \midrule
         \rowcolor{synthetic_c!25} \multicolumn{8}{c}{\textit{\textbf{Video Understanding}}} \\
         Video-MME & \citep{fu2025video} & Video & 40 & 40 & 40 & 40 & 160\\
         LongVideoBench & \citep{wu2024longvideobench} & Video & 40 & 40 & 40 & 40 & 160\\
         \bottomrule
    \end{tabular}}
    
    \label{tab:data_statistic}
\end{table*}

%% file: section/experiment.tex
\section{Experiment}

\subsection{Experimental Setup}
We conduct a comprehensive evaluation of 28 LCVLMs, comprising 22 open-source\footnote{We exclude InternVL3.5-4B and the Seed-2.0 series from the visual grounding task for different reasons: InternVL3.5-4B's specific ViT mechanism yields context lengths exceeding 600K in multi-image scenarios, surpassing our available computational resources, while the Seed-2.0 series exceed its own context window under this setting.} and 6 proprietary models\footnote{Claude and GPT series are not evaluated because their restrictive API limits (100 images per request for Claude, 50 for GPT) fall short of the input requirements for MMLongCite.}, as detailed in Table~\ref{tab:model_statistic}.
The evaluated models exhibit considerable diversity in their specifications: context windows (40K to 1M tokens), model parameter (2B to 235B), architectures (including both Dense and MoE designs) and diverse reasoning paradigms (i.e., non-thinking, thinking, and hybrid modes), covering a broad spectrum of model scales and architectures.

Following  ~\citep{gao2023enabling,tang2025citeeval,zhang2025longcite}, we evaluate our model on two key aspects: citation quality and generation quality. To measure citation quality, we report citation precision ($CP$), recall ($CR$), and the resulting citation $F_1$ ($F_1$). To evaluate generation quality, we measure correctness ($\mathbf{Cor}$) by scoring each response on a 0/0.5/1 scale (incorrect/partially correct/fully correct) against the ground truth. For both evaluation aspects, we leverage GPT-5.2\footnote{https://openai.com/index/introducing-gpt-5-2/}~\citep{gpt_5_2_2025} as the adjudicator to ensure higher accuracy, compared to prior work~\citep{gao2023enabling,tang2025citeeval}. Detailed definitions of these evaluation metrics are provided in Appendix~\ref{appendix:metrics}, while a discussion on the reliability of employing LLM-as-a-judge is presented in Appendix~\ref{appendix:human_evaluation}.

\input{table/main_cite}

\input{table/hard_cite}

\subsection{Main Results}
We present the evaluation results on \textsc{MMLongCite} in Table~\ref{tab:main_cite} and the model performance on \textsc{MMLongCite-HR} in Table~\ref{tab:hard_cite}.

\subsubsection{Analysis of MMLongCite}

\paragraph{Performance of Open-source Models}
Our investigation yields three key insights. 1) We observe that scaling within a well-designed model family yields consistent gains in citation quality. The Qwen2.5-VL and Qwen3-VL series both exhibit a clear monotonic trend where citation $F_1$ improves as parameters increase. This confirms that model scaling remains an effective strategy for enhancing citation capability. 2) Beyond scaling, training recipe is equally pivotal: medium-sized models such as Qwen3.5-4B already surpass the 72B-parameter Qwen2.5-VL-72B, underscoring the role of data quality and training strategies amidst the parameter scaling race~\citep{coreteam2025mimovltechnicalreport}. 3) Our results reveal divergent model capabilities across modalities. For instance, the Gemma-4 series secures standout citation $F_1$ scores in Video Understanding, defying its overall ranking on other tasks. This suggests certain models possess domain-specific strengths not captured by broad-spectrum benchmarks.

\paragraph{Open-source Models vs. Proprietary Models}
A significant performance gap persists between open-source and proprietary models, particularly in retrieval-intensive tasks. This disparity is most pronounced on Visual Grounding, which requires LCVLMs to precisely retrieve fine-grained evidence from long contexts. On this task, the leading open-source model (Qwen3.5-122B-A10B) still trails the top proprietary model (Gemini-3-Flash) by nearly 12 points. A similar gap is observed on Video Understanding, where Gemini-3-Pro leads by a comparable margin. Furthermore, proprietary models tend to achieve a more balanced trade-off between citation precision and recall, whereas open-source models—even those exceeding 100B parameters—generally exhibit a recall-skewed deficit.

\paragraph{Impact of Reasoning Mode}
The comparison between models with and without reasoning mode highlights that CoT acts as a double-edged sword in generating answers with citations. As evidenced by MiMo-VL-7B-RL, enabling the reasoning mode consistently boosts answer correctness and citation precision, yet it incurs a marked decline in citation recall. A similar pattern is observed for Kimi-VL-A3B and MiMo-VL-7B-SFT. This suggests a fundamental trade-off: the CoT mechanism promotes more deliberative reasoning, boosting the quality of the final answer and the precision of its cited evidence, but it also induces a conservative citation behavior, causing the model to overlook relevant evidence. Notably, GLM-4.6V partially escapes this trade-off, with its think mode improving both precision and recall on most tasks, suggesting that the trade-off is modulated by the underlying training recipe rather than being intrinsic to CoT itself.

\paragraph{Correlation Between Citation Quality and Correctness}
We identify a decoupling of citation quality and answer correctness in small models (i.e., models with 7B parameters or fewer). For instance, Qwen2.5-VL-7B-Ins achieves 51.23 in correctness on Visual Grounding but only 17.92 in citation $F_1$; a similar gap is observed for Qwen3-VL-2B-Ins. This indicates that small models are prone to generating hallucinated citations rather than grounding their responses in the provided context, which is consistent with the observations of \citep{tang2025citeeval} and \citep{small_hallu}. The phenomenon underscores that correctness alone is an insufficient metric for evaluating LCVLMs.

\subsubsection{Analysis of MMLongCite-HR}
To evaluate the robustness of models when handling high-resolution inputs with varying information densities, we report the performance on \textsc{MMLongCite-HR} in Table~\ref{tab:hard_cite}. It introduces a "grid-stitching" paradigm, where visual contexts are aggregated into high-resolution images—4 images per grid for the "Easy" setting and 16 for the "Hard" setting. The shaded rows represent the standard \textsc{MMLongCite} performance for reference.

\paragraph{Sensitivity to Visual Information Density.}
The most prominent observation is the pervasive performance degradation when transitioning from the standard sequence input to the high-resolution grid input. Although the grid-stitching operation preserves the original resolution, it fundamentally alters the spatial locality from the perspective of the Vision Transformer. By serializing a larger grid into a 1D sequence, adjacent patches within a single sub-image are pushed significantly apart in the token sequence. This disruption of spatial adjacency imposes a substantial challenge for the attention mechanism to reconstruct local visual semantics and perform accurate citation across a prolonged context. This impact is particularly severe in fine-grained tasks such as LongDocURL. For example, Qwen2.5-VL-7B-Ins sees its citation $F_1$ plummet from 25.69 in the standard setting to 15.19 in the Easy setting. Furthermore, as complexity scales from Easy to Hard, most models exhibit additional declines; the $F_1$ score of Gemini-3-Pro on LongDocURL drops significantly from 84.68 to 63.78. These trends suggest that current LCVLMs, despite claiming high-resolution support, struggle with the spatial entanglement and densed feature resolution of individual documents when densely packed into a single composite view, impairing their ability to precisely localize and cite evidence.

\paragraph{Architecture and Scale as Determinants of Robustness.}
While all models are affected by the high resolution shift, the results highlight a stark dichotomy in robustness correlating with model scale and nature (proprietary vs. open-source). Gemini-3-Pro exhibits superior resilience, maintaining a commanding lead across all benchmarks; even in the Hard setting of 2WikiMultiHopQA, it achieves a citation $F_1$ of 76.77, vastly outperforming the largest open-source contender, Qwen2.5-VL-72B-Ins, which scores only 20.39. Among the open-source models, scaling laws hold firm: the 72B parameter model consistently outperforms its 7B counterpart, indicating that larger language backbones are better equipped to disambiguate the noisy visual signals generated by dense image grids. Interestingly, Qwen3-VL-30B-A3B-Ins demonstrates competitive performance in multi-hop reasoning tasks despite being smaller than the 72B model, suggesting that algorithmic improvements in the "active" parameters or architecture may offer a pathway to efficiency without sacrificing grounding capabilities in complex visual scenarios.

%% file: table/main_cite.tex
\begin{table*}[t]
    \centering
    \caption{Performance of LCVLMs in \textsc{MMLongCite}.}
    \resizebox{\textwidth}{!}{
    \begin{tabular}{l | cccc | cccc | cccc | cccc}
        \toprule
         \multirow{2}{*}{\bf Models} &  \multicolumn{4}{c|}{\bf Single-Source Visual QA} &  \multicolumn{4}{c|}{\bf Multi-Source Visual QA} &  \multicolumn{4}{c|}{\bf Visual Grounding } &  \multicolumn{4}{c}{\bf Video Understanding } \\
         \cmidrule{2-17}
         & \makecell[c]{$\mathbf{CP}$} & \makecell[c]{\bf $\mathbf{CR}$} & \makecell[c]{\bf $\mathbf{F_1}$}  & \makecell[c]{\bf $\mathbf{Cor}$}  &\makecell[c]{$\mathbf{CP}$} & \makecell[c]{\bf $\mathbf{CR}$} & \makecell[c]{\bf $\mathbf{F_1}$}  & \makecell[c]{\bf $\mathbf{Cor}$}  & \makecell[c]{$\mathbf{CP}$} & \makecell[c]{\bf $\mathbf{CR}$} & \makecell[c]{\bf $\mathbf{F_1}$}  & \makecell[c]{\bf $\mathbf{Cor}$}  & \makecell[c]{$\mathbf{CP}$} & \makecell[c]{\bf $\mathbf{CR}$} & \makecell[c]{\bf $\mathbf{F_1}$} & \makecell[c]{\bf $\mathbf{Cor}$}   \\
         \midrule
          \rowcolor{gray!20} \multicolumn{17}{c}{\textit{\textbf{Proprietary Models}}}  \\
         Gemini-3-Pro(think) & \bf 84.49 & \underline{83.58} & \bf 83.36 & \underline{80.86} & \underline{83.75} & \underline{84.97} & \underline{84.02} & \bf 85.85& \bf 80.48 & \underline{72.54} & \underline{75.07} & \bf 87.28& \bf 77.97 & \underline{69.39} & \bf 71.75 & \bf 85.31 \\
Gemini-3-Flash(think) & \underline{79.61} & \bf 83.96 & \underline{81.06} & \bf 82.73& \bf 86.22 & \bf 89.13 & \bf 87.22 & \underline{85.40} & \underline{78.54} & \bf 86.29 & \bf 81.32 & \underline{84.52} & 72.40 & \bf 74.24 & \underline{71.74} & \underline{84.69} \\
Seed-2.0-Pro & 76.26 & 68.38 & 70.57 & 73.67& 81.70 & 74.34 & 76.56 & 78.71& - & - & - & -& \underline{73.14} & 46.34 & 54.63 & 82.87 \\
Seed-2.0-Lite & 53.41 & 51.50 & 51.22 & 71.30& 65.78 & 68.41 & 66.08 & 84.71& - & - & - & -& 50.74 & 40.33 & 42.72 & 79.72 \\
Seed-2.0-Mini & 68.73 & 56.49 & 59.10 & 69.90& 69.66 & 65.14 & 64.92 & 76.84& - & - & - & -& 67.49 & 38.06 & 46.67 & 78.77 \\
Seed-1.6 & 64.39 & 63.57 & 62.86 & 63.67& 76.79 & 77.77 & 75.89 & 78.00& 72.92 & 70.50 & 71.29 & 72.84& 67.56 & 67.53 & 64.79 & 80.16 \\
\rowcolor{gray!20} \multicolumn{17}{c}{\textit{\textbf{Open-source Models}}}  \\
Qwen2.5-VL-3B-Ins & 5.90 & 6.97 & 5.91 & 35.23& 9.36 & 11.80 & 9.86 & 40.17& 4.75 & 3.65 & 3.93 & 43.02& 22.71 & 25.56 & 22.50 & 66.25 \\
Qwen2.5-VL-7B-Ins & 28.80 & 32.21 & 28.63 & 46.09& 33.24 & 36.60 & 33.10 & 52.59& 19.67 & 18.10 & 17.92 & 51.23& 41.30 & 42.07 & 38.89 & 68.44 \\
Qwen2.5-VL-32B-Ins & 47.49 & 34.33 & 38.00 & 55.62& 51.23 & 38.56 & 42.12 & 63.81& 43.83 & 28.20 & 32.98 & 62.45& 51.46 & 26.86 & 33.33 & 69.69 \\
Qwen2.5-VL-72B-Ins & 57.80 & 56.33 & 56.31 & 59.92& 59.99 & 60.94 & 59.77 & 60.37& 54.47 & 46.55 & 49.07 & 61.81& 66.15 & 52.60 & 56.03 & 72.97 \\
Qwen3-VL-2B-Ins & 26.67 & 26.35 & 24.83 & 40.78& 29.55 & 33.71 & 29.94 & 45.73& 29.56 & 27.73 & 27.56 & 60.56& 53.31 & 38.94 & 43.14 & 64.69 \\
Qwen3-VL-4B-Ins & 53.56 & 48.92 & 50.26 & 52.11& 57.74 & 54.41 & 55.16 & 58.47& 49.10 & 35.26 & 39.61 & 57.25& 68.42 & 53.70 & 57.73 & 71.88 \\
Qwen3-VL-8B-Ins & 61.58 & 53.33 & 55.86 & 59.77& 69.40 & 58.66 & 61.88 & 65.56& 42.10 & 30.22 & 33.75 & 57.55& 70.67 & 60.70 & 62.71 & 68.91 \\
Qwen3-VL-30B-A3B-Ins & 64.50 & 57.70 & 59.47 & 60.47& 74.15 & 67.03 & 68.91 & 67.65& 57.00 & 44.12 & 48.15 & 73.33& 66.71 & 43.80 & 49.92 & 76.41 \\
Qwen3-VL-235B-A22B-Ins & 69.16 & \underline{65.25} & \underline{65.98} & 68.36& 77.86 & \underline{76.58} & \underline{75.78} & 74.55& \underline{73.68} & \underline{61.71} & \underline{65.34} & 66.33& 69.93 & 62.83 & 63.84 & \bf 80.62 \\
MiMo-VL-7B-SFT & 48.56 & 44.52 & 45.07 & 54.30& 62.70 & 60.95 & 60.39 & 68.49& 43.34 & 39.93 & 40.47 & 54.91& 55.21 & 49.14 & 49.57 & 70.62 \\
MiMo-VL-7B-SFT(think) & 47.30 & 35.54 & 38.70 & 59.22& 63.45 & 53.77 & 56.00 & 72.51& 40.92 & 34.26 & 36.04 & 53.61& 50.41 & 24.93 & 31.46 & 73.44 \\
MiMo-VL-7B-RL & 55.69 & 45.95 & 48.66 & 55.31& 56.42 & 52.75 & 52.83 & 61.63& 45.25 & 41.03 & 41.64 & 55.17& 56.07 & 44.81 & 47.59 & 72.34 \\
MiMo-VL-7B-RL(think) & 55.17 & 33.08 & 39.20 & 61.64& 59.39 & 44.62 & 48.57 & 67.83& 56.78 & 39.71 & 44.29 & 55.95& 49.32 & 22.29 & 28.60 & 73.44 \\
Kimi-VL-A3B-Ins & 28.84 & 26.07 & 26.10 & 33.59& 44.37 & 37.69 & 39.07 & 54.85& 23.91 & 21.78 & 22.42 & 62.31& 20.32 & 16.45 & 17.06 & 67.81 \\
Kimi-VL-A3B-Think & 44.64 & 28.60 & 32.67 & 45.39& 50.79 & 34.93 & 38.83 & 46.85& 28.64 & 21.40 & 23.47 & 42.02& 39.18 & 15.78 & 21.04 & 55.00 \\
InternVL3\_5-4B & 14.48 & 11.35 & 11.99 & 23.28& 12.98 & 11.42 & 11.68 & 20.18& - & - & - & -& 15.07 & 13.97 & 13.76 & 43.75 \\
Gemma-4-E4B & 42.60 & 46.63 & 43.99 & 38.28& 66.55 & 70.51 & 67.88 & 42.83& 44.61 & 47.03 & 45.28 & 54.34& 64.52 & \bf 68.74 & \underline{64.99} & 45.47 \\
Gemma-4-26B-A4B & 48.77 & 49.28 & 48.48 & 45.70& 64.71 & 66.60 & 65.33 & 51.40& 41.16 & 43.98 & 42.03 & 59.97& 68.54 & \underline{66.38} & \bf 65.34 & 69.84 \\
GLM-4.6V-Flash & 56.40 & 43.43 & 46.71 & 59.53& 59.34 & 52.25 & 53.06 & 64.67& 53.84 & 45.94 & 48.31 & 77.58& 48.54 & 24.58 & 30.78 & 74.22 \\
GLM-4.6V & 58.76 & 53.06 & 53.99 & 64.53& 66.47 & 65.06 & 63.69 & 72.14& 46.04 & 41.06 & 42.64 & \underline{78.12}& 49.60 & 50.75 & 48.23 & 76.41 \\
GLM-4.6V(think) & 62.80 & 58.51 & 58.99 & \underline{70.94} & 69.15 & 71.99 & 68.19 & \underline{75.36} & 54.01 & 50.80 & 51.54 & \bf 80.12& 61.23 & 52.53 & 53.42 & 78.28 \\
Qwen3.5-2B & 44.33 & 31.55 & 35.16 & 48.98& 47.86 & 38.37 & 40.80 & 56.70& 36.51 & 27.63 & 30.21 & 66.77& 54.54 & 29.19 & 35.94 & 71.88 \\
Qwen3.5-4B & 70.12 & 60.36 & 63.50 & 63.91& 79.62 & 69.06 & 72.60 & 69.07& 63.47 & 52.92 & 55.87 & 58.03& \underline{70.89} & 48.20 & 54.71 & 77.19 \\
Qwen3.5-9B & \underline{70.52} & 53.29 & 58.14 & 69.61 & \underline{80.69} & 66.80 & 71.14 & 73.98& 65.50 & 52.65 & 56.30 & 68.61& 61.31 & 48.77 & 51.54 & 77.34 \\
Qwen3.5-122B-A10B & \bf 77.83 & \bf 69.06 & \bf 71.33 & \bf 71.95& \bf 85.10 & \bf 78.69 & \bf 80.51 & \bf 78.26& \bf 78.42 & \bf 62.36 & \bf 67.36 & 75.45&  \bf 73.27 &  50.87 & 56.87 &  \underline{80.16} \\

         \bottomrule
    \end{tabular}}
    \label{tab:main_cite}
\end{table*}

%% file: table/hard_cite.tex
\begin{table*}[t]
    \centering
    \caption{Performance of LCVLMs on \textsc{MMLongCite-HR}. For each model, the gray-shaded row reports its counterpart results on \textsc{MMLongCite}, included for cross-setting comparison.}
    \resizebox{\textwidth}{!}{
    \begin{tabular}{l | cc | cc | cc | cc | cc | cc | cc | cc}
        \toprule
         \multirow{3}{*}{\bf Models} &  \multicolumn{4}{c|}{\bf LongDocURL } &  \multicolumn{4}{c|}{\bf 2WikiMultiHopQA} &  \multicolumn{4}{c|}{\bf Visual Haystack } &  \multicolumn{4}{c}{\bf LongVideoBench } \\
         \cmidrule{2-17}
         & \multicolumn{2}{c|}{\bf $\mathbf{F_1}$} & \multicolumn{2}{c|}{\bf $\mathbf{Cor}$} & \multicolumn{2}{c|}{\bf $\mathbf{F_1}$} & \multicolumn{2}{c|}{\bf $\mathbf{Cor}$} & \multicolumn{2}{c|}{\bf $\mathbf{F_1}$} & \multicolumn{2}{c|}{\bf $\mathbf{Cor}$} & \multicolumn{2}{c|}{\bf $\mathbf{F_1}$} & \multicolumn{2}{c}{\bf $\mathbf{Cor}$} \\
         \cmidrule{2-3} \cmidrule{4-5} \cmidrule{6-7} \cmidrule{8-9} \cmidrule{10-11} \cmidrule{12-13} \cmidrule{14-15} \cmidrule{16-17}
         & \bf Easy & \bf Hard & \bf Easy & \bf Hard & \bf Easy & \bf Hard & \bf Easy & \bf Hard & \bf Easy & \bf Hard & \bf Easy & \bf Hard & \bf Easy & \bf Hard & \bf Easy & \bf Hard   \\
         \midrule
         \multirow{2}{*}{Gemini-3-Pro} & \multicolumn{2}{c|}{\cellcolor{gray!20}87.07} & \multicolumn{2}{c|}{\cellcolor{gray!20}85.78} & \multicolumn{2}{c|}{\cellcolor{gray!20}76.68} & \multicolumn{2}{c|}{\cellcolor{gray!20}83.12} & \multicolumn{2}{c|}{\cellcolor{gray!20}78.08} & \multicolumn{2}{c|}{\cellcolor{gray!20}76.75} & \multicolumn{2}{c|}{\cellcolor{gray!20}67.56} & \multicolumn{2}{c}{\cellcolor{gray!20}80.62} \\
         & 84.68 & 63.78 & 85.31 & 70.31 & 78.17 & 76.77 & 83.75 & 85.47 & 77.11 & 74.81 & 74.12 & 62.75 & 65.02 & 67.51 & 79.06 & 75.62 \\
         \midrule
         \multirow{2}{*}{Qwen2.5-VL-7B-Ins} & \multicolumn{2}{c|}{\cellcolor{gray!20}25.69} & \multicolumn{2}{c|}{\cellcolor{gray!20}49.06} & \multicolumn{2}{c|}{\cellcolor{gray!20}21.18} & \multicolumn{2}{c|}{\cellcolor{gray!20}41.25} & \multicolumn{2}{c|}{\cellcolor{gray!20}28.08} & \multicolumn{2}{c|}{\cellcolor{gray!20}52.62} & \multicolumn{2}{c|}{\cellcolor{gray!20}36.92} & \multicolumn{2}{c}{\cellcolor{gray!20}65.31} \\
         & 15.19 & 12.81 & 50.00 & 45.62 & 11.25 & 13.22 & 38.44 & 38.91 & 21.38 & 26.15 & 54.00 & 53.12 & 30.12 & 26.34 & 61.25 & 61.25  \\
         \midrule
          \multirow{2}{*}{Qwen2.5-VL-72B-Ins} & \multicolumn{2}{c|}{\cellcolor{gray!20}56.87} & \multicolumn{2}{c|}{\cellcolor{gray!20}65.00} & \multicolumn{2}{c|}{\cellcolor{gray!20}42.20} & \multicolumn{2}{c|}{\cellcolor{gray!20}42.34} & \multicolumn{2}{c|}{\cellcolor{gray!20}53.27} & \multicolumn{2}{c|}{\cellcolor{gray!20}48.00} & \multicolumn{2}{c|}{\cellcolor{gray!20}52.67} & \multicolumn{2}{c}{\cellcolor{gray!20}64.38} \\
         & 33.72 & 20.70 & 61.25 & 54.84 & 20.41 & 20.39 & 40.00 & 36.41 & 52.96 & 51.84 & 43.88 & 41.88 & 48.17 & 46.42 & 60.94 & 65.00   \\
         \midrule
           \multirow{2}{*}{Qwen3-VL-30B-A3B-Ins} & \multicolumn{2}{c|}{\cellcolor{gray!20}57.44} & \multicolumn{2}{c|}{\cellcolor{gray!20}63.12} & \multicolumn{2}{c|}{\cellcolor{gray!20}59.36} & \multicolumn{2}{c|}{\cellcolor{gray!20}61.72} & \multicolumn{2}{c|}{\cellcolor{gray!20}51.05} & \multicolumn{2}{c|}{\cellcolor{gray!20}60.25} & \multicolumn{2}{c|}{\cellcolor{gray!20}51.44} & \multicolumn{2}{c}{\cellcolor{gray!20}73.75} \\
         & 26.41 & 30.41 & 59.53 & 55.00 & 19.04 & 33.78 & 54.84 & 56.09 & 41.94 & 38.95 & 51.00 & 42.12 & 46.76 & 44.32 & 70.94 & 51.25  \\
         \bottomrule
    \end{tabular}}
    \label{tab:hard_cite}
\end{table*}

%% file: section/analysis.tex
\begin{figure}[t]
    \centering
    \includegraphics[width=\linewidth]{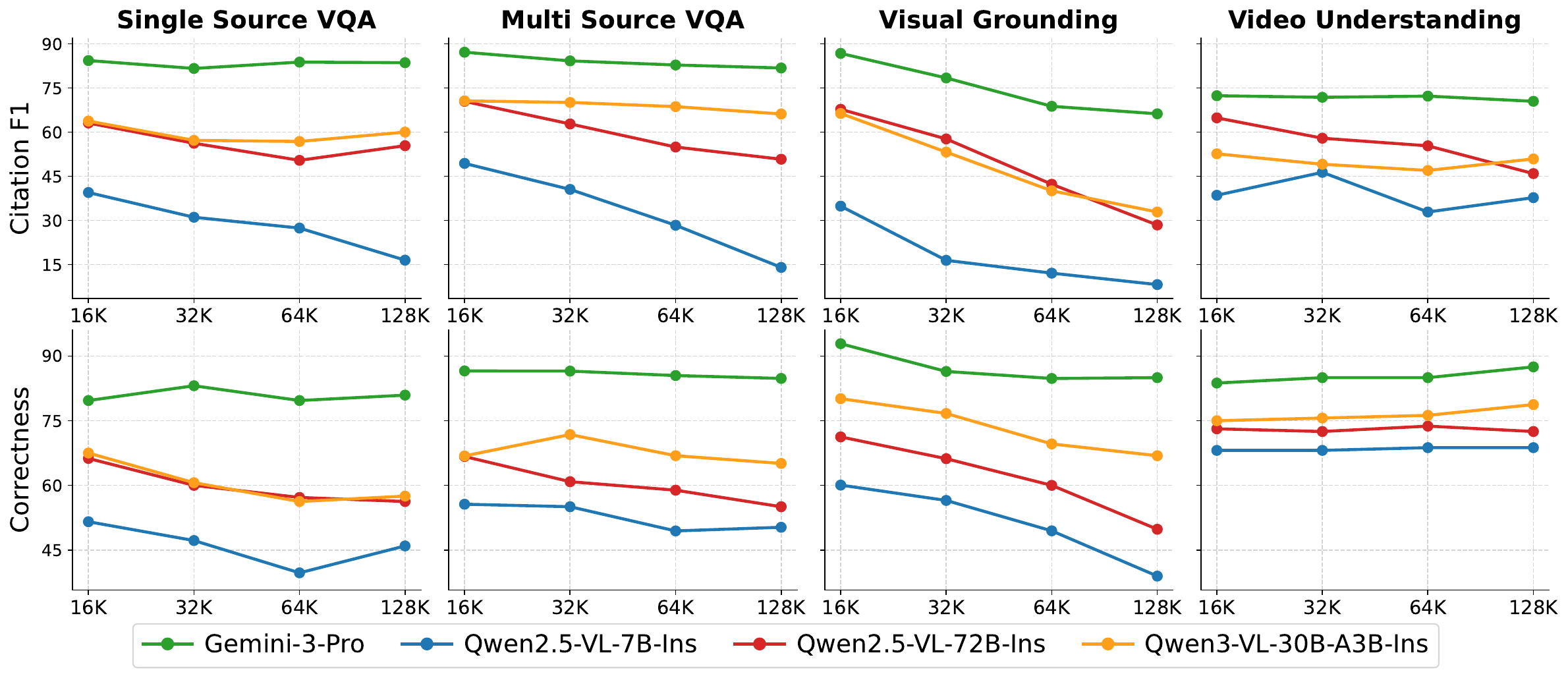}
    \caption{Model Performance across different context length intervals.}
    \label{fig:length_exp}
\end{figure}

\begin{figure*}[t]
    \centering
    \includegraphics[width=\linewidth]{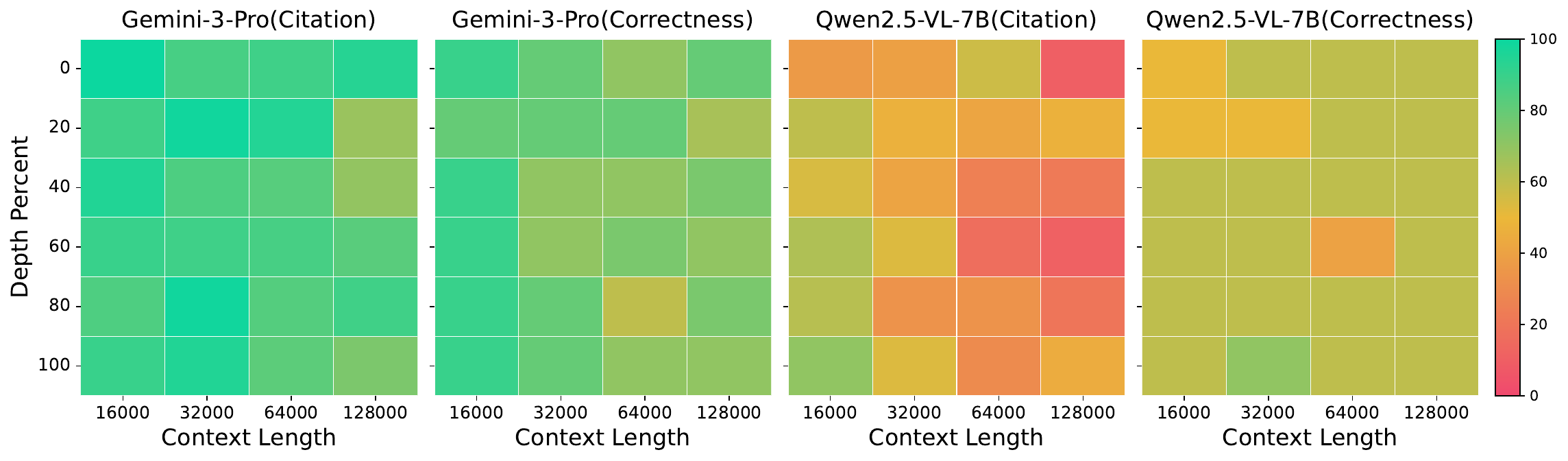}
    \caption{Model Performance on citation quality and correctness across different depths and context lengths in Visual Haystack.}
    \label{fig:needle}
\end{figure*}

\begin{figure}[t]
    \centering
    \includegraphics[width=\linewidth]{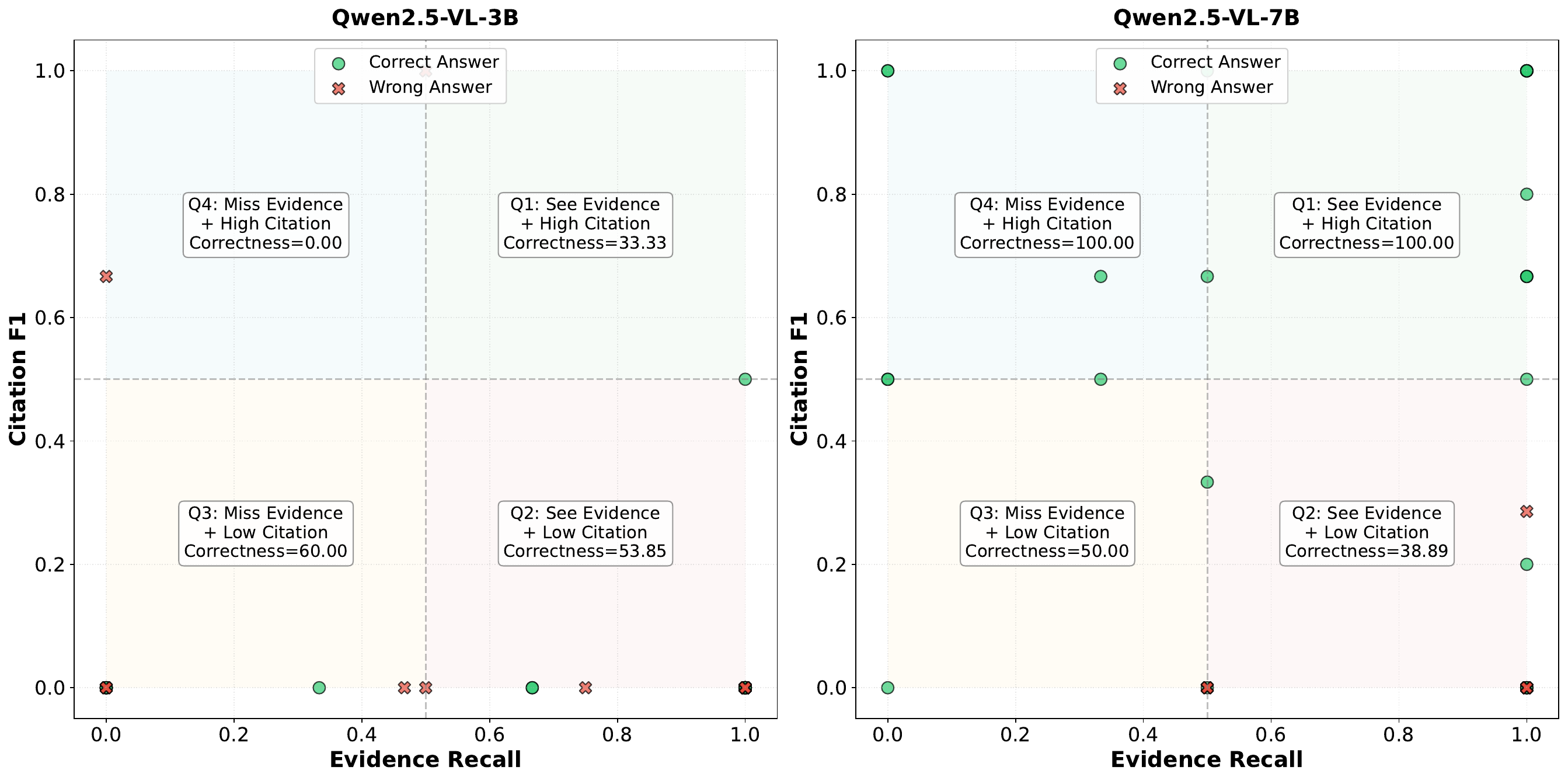}
    \caption{Correlation between attention patterns, citation quality, and model correctness.}
    \label{fig:fail_mode}
\end{figure}

\section{Analysis}
In this section, we conduct a series of analyses to further probe the model's capabilities in handling multimodal long contexts. Specifically, we investigate three key aspects: (1) the impact of varying context lengths on the faithfulness and correctness of models in~\cref{ablation:length_interval}; (2) the performance variation of LCVLMs with respect to the position of key information within the context in~\cref{ablation:needle_pos}; (3) the relationship between attention patterns and failure modes in long contexts in~\cref{ablation:attn_pattern}.

\subsection{Analysis of different length intervals}
\label{ablation:length_interval}

Figure~\ref{fig:length_exp} dissects model behavior across four length intervals on a per-task basis, revealing degradation patterns that are markedly task-dependent. We distill three salient observations. First, the proprietary–open-source gap widens with task complexity rather than length alone: Gemini-3-Pro maintains near-flat trajectories across every task, whereas open-source models erode sharply on tasks demanding precise evidence localization. Second, visual grounding emerges as the most fragile regime, with all open-source models suffering monotonic, steep declines in citation $F_1$. We attribute this to the finer granularity of grounding inputs—each natural image consumes far fewer tokens than a document page or video frame, so an identical token budget yields a substantially larger candidate pool, inflating the index space over which models must disambiguate. Video understanding, by contrast, remains remarkably stable owing to temporal redundancy and coarse-grained semantic cues that ease pinpoint localization. Third, citation faithfulness deteriorates substantially faster than answer correctness across nearly all tasks, indicating that the ability to localize supporting evidence degrades more rapidly with context length than the ability to produce a plausible answer, and underscoring faithfulness as an orthogonal and more demanding axis for long-context evaluation.

\subsection{Analysis of the Position of Needles}
\label{ablation:needle_pos}

We investigate the impact of needle position and context length on model performance in Visual Haystack, with results visualized in Figure~\ref{fig:needle}. While Gemini-3-Pro exhibits near-perfect resilience across all depths and lengths, Qwen2.5-VL-7B-Ins reveals a critical decoupling between retrieval and attribution. Despite a nominal 128K context window, its effective window for citation significantly lags behind correctness~\citep{an2024does}. Crucially, the model suffers from a pronounced 'lost-in-the-middle' phenomenon~\citep{liu2024lost}—performance collapses in central depths (40\%–80\%) during source localization—exposing precise positional locating as a severe bottleneck for open-source models.

\subsection{Correlation between Attention Patterns and Failure Modes}
\label{ablation:attn_pattern}

We conduct a fine-grained analysis to investigate whether failures stem from missing the evidence (``Failure to See'') or mishandling it (``Failure to Use'') in Figure~\ref{fig:fail_mode}. Specifically, we calculate the proportion of evidence that received the high attention score of models during generation as Evidence Recall. By mapping Evidence Recall against citation $F_1$, we observe that for Qwen2.5-VL-7B-Ins, high attention is necessary but insufficient for correctness: the model achieves high accuracy when both metrics are high, but drops to 38.89 in the ``See Evidence + Low Citation'' regime. This discrepancy indicates that a primary failure mode is not a lack of retrieval, but the inability to effectively utilize the attended information. Furthermore, smaller models like Qwen2.5-VL-3B-Ins show low correctness even in the ``See Evidence + High Citation'' quadrant, suggesting their limitation lies in fundamental comprehension rather than attention allocation. More details can be found in Appendix~\ref{appendix:attention} and we also conduct a fine-grained error analysis in Appendix~\ref{appendix:error_analysis}.

%% file: section/conclusion.tex
\section{Conclusion}

In this paper, we introduce \textsc{MMLongCite}, a comprehensive benchmark specifically designed to evaluate the faithfulness of long-context vision-language models. The benchmark encompasses 4 major task categories spanning 8 multimodal datasets, with context lengths scaling from 16K to 128K.
Evaluating 28 cutting-edge LCVLMs reveals a severe decoupling between answer correctness and citation faithfulness. \textsc{MMLongCite-HR} further exposes the fragility of models in isolating visual evidence within high-resolution spaces. Additionally, we find that advanced reasoning strategies present a trade-off between generation performance and citation recall. Finally, our mechanistic analysis uncovers a discrepancy between attention allocation and information utilization, highlighting a fundamental ``failure to use'' bottleneck in LCVLMs.
We hope our benchmark and findings will inspire future research to develop more reliable, context-faithful LCVLMs capable of handling complex, real-world multimodal streams.

%% file: section/appendix.tex
\appendix

\clearpage
\input{table/model_statistics}
\section{Details of LCVLMs}
\label{appendix:models}
We list the detailed information of the models evaluated in this paper in Table~\ref{tab:model_statistic}. All experiments are done on 8$\times$H20 GPUs.

\section{Evaluation Settings}
\label{appendix:metrics}

To holistically assess the model performance, we assess both the quality of the citations and the factual correctness of the generated content. Our evaluation methodology leverages a powerful LCVLM, GPT-5.2, as an automated judge for all metrics.

\subsection{Citation Quality}

We first evaluate the quality of citations in the generated responses using three metrics: Citation Recall (\texttt{CR}), Citation Precision (\texttt{CP}), and their harmonic mean, Citation F1. For each generated statement, the judge determines if the cited sources provide factual support. The sources in our multimodal context can be either text segments or images. We present the prompt for evaluating citation quality in Figure~\ref{fig:gpt4.1_evaluate_citation}.

\paragraph{Citation Recall (\texttt{CR})}
This metric measures whether the complete set of cited sources for a given statement provides sufficient factual support. For a response containing a set of statements $S$, we gather all cited sources $C(s)$ for each statement $s \in S$. The recall is then calculated as the fraction of statements that are fully supported by their cited sources.
\begin{equation}
    \texttt{CR} = \frac{1}{|S|} \sum_{s \in S} \mathbb{I}(f_{\text{eval}}(C(s), s) = \text{relevant})
\end{equation}
where $\mathbb{I}(\cdot)$ is the indicator function and $f_{\text{eval}}$ is the LCVLM-based evaluator. A score of 1 indicates that the collection of sources $C(s)$ factually entails the statement $s$.

\paragraph{Citation Precision (\texttt{CP})}
This metric assesses the necessity of each individual citation, penalizing superfluous or irrelevant sources. We evaluate precision using a leave-one-out approach. A citation is deemed necessary only if the full set of sources supports the statement, but the set with this specific citation removed does not.
\begin{equation}
    \texttt{CP} = \frac{\sum_{s \in S} \sum_{c \in C(s)} \mathbb{I}_{\text{necessary}}(c, s)}{|\mathcal{C}_{\text{total}}|}
\end{equation}
where $\mathcal{C}_{\text{total}}$ is the total number of citations in the response. The indicator $\mathbb{I}_{\text{necessary}}(c, s)$ is 1 if and only if both of the following conditions are met:
\begin{itemize}
    \item The complete set of sources supports the statement: $f_{\text{eval}}(C(s), s) = \text{entailment}$.
    \item The set of sources without the specific citation $c$ fails to support the statement: $f_{\text{eval}}(C(s) \setminus \{c\}, s) \neq \text{entailment}$.
\end{itemize}

\paragraph{Citation $\mathbf{F_1}$ (\texttt{$F_1$})}
We compute the F1-score as the harmonic mean of Citation Recall and Citation Precision, providing a single, comprehensive measure of overall citation quality.
\begin{equation}
    \texttt{$F_1$} = \frac{2 \times \texttt{CR} \times \texttt{CP}}{\texttt{CR} + \texttt{CP}}
\end{equation}

\subsection{Generation Quality}

\paragraph{Correctness (\texttt{Cor})}
Beyond citation quality, we assess factual correctness by benchmarking the response against a human-authored ground truth. The LCVLM judge initially assigns an integer score on a scale of 1 to 3, which is then mapped to the discrete set \{0, 0.5, 1\}. We present the prompt for evaluating correctness in Figure~\ref{fig:gpt4.1_evaluate_correctness}. The criteria are as follows:
\begin{itemize}
    \item \textbf{1 (Fully Correct):} The generated response accurately and completely covers the information present in the ground-truth answer. It contains no factual errors or significant omissions.
    \item \textbf{0.5 (Partially Correct):} The response is broadly correct but suffers from minor factual inaccuracies, hallucinations, or omits key information found in the ground-truth answer.
    \item \textbf{0 (Incorrect):} The response contains major factual errors, is irrelevant to the question, or is nonsensical.
\end{itemize}
The final correctness score is the average score over all instances in the test set. For a dataset with $N$ instances, the score is calculated as:
\begin{equation}
    \texttt{Cor} = \frac{1}{N} \sum_{i=1}^{N} \text{Score}(R_i, A_i)
\end{equation}
where $R_i$ is the model-generated response for the $i$-th instance, $A_i$ is the corresponding ground-truth answer, and the $\text{Score}(\cdot, \cdot)$ function represents the score assigned by the LCVLM judge.

\input{table/full_cite1}
\input{table/full_cite2}

\section{Details of Benchmark Construction}
\label{appendix:benchmark_construction}
In this section, we provide a granular description of the data processing and construction pipeline for the datasets included in our benchmark. Our primary goal is to ensure a balanced distribution across the four length intervals (0-16K, 16-32K, 32-64K, and 64-128K tokens) while maintaining the integrity of the evaluation tasks.

For datasets focusing on long document understanding, namely LongDocURL and MMLongBench-Doc, each instance comprises one or several evidence pages essential for answering the query. A limitation of the original LongDocURL dataset is its preprocessing step, which truncates the input to approximately 30 pages surrounding the evidence pages. This truncation results in insufficient data volume to support our evaluation across the longer context intervals (specifically the 32-64k and 64-128K ranges). To address this, we revert to the raw documents from the original corpus rather than the truncated versions. We then re-process these documents following the truncation strategy described in the main text. This approach allows us to strictly control the context length while preserving the position bias of the evidence set, ensuring a rigorous evaluation of long-context capabilities.

SlideVQA and 2WikiMultiHopQA are employed to evaluate the model's multi-hop reasoning capabilities over long contexts composed of multiple distinct documents. Since 2WikiMultiHopQA is originally a text-only dataset, we first perform a modality alignment step using the Pillow\footnote{https://pillow.readthedocs.io/en/stable/} library to render each paragraph into a visual document image, unifying the input format across tasks. We then construct distractor sets to extend the context length. For SlideVQA, we treat each slide deck (comprising 20 pages) as a cohesive unit for distraction; similarly, for 2WikiMultiHopQA, each paragraph is treated as an individual unit. Following the methodology outlined in the main paper, we randomly sample distractors to pad the (Question, Answer, Evidence) triplets to the target length intervals.

Unlike other tasks in our benchmark that rely on inherently cohesive documents or pre-curated noise pools, multi-source visual QA tasks (SlideVQA and 2WikiMultiHopQA) require sampling distractors from external sources across the corpus to extend the context. This external sampling introduces a unique risk of distractor collision—where randomly sampled noise images might inadvertently contain alternative evidence or conflicting information. To ensure the rigor of our evaluation and mitigate this risk, we implement a strict LVLM-based filtering pipeline. Specifically, for each candidate distractor image, we utilize a powerful open-source model (Qwen/Qwen3.5-397B-A17B) to perform a logical consistency check. The model is prompted to act as a strict evaluator, inspecting whether the visual information within the candidate distractor could contradict the provided ground truth. Any distractor flagged as presenting an alternate-evidence trap is explicitly removed from the candidate pool. This rigorous curation guarantees that the distractors serve exclusively as noise to scale the context length, effectively preventing the introduction of factual conflicts or ambiguous clues that could compromise the uniqueness and absolute correctness of the ground truth.

The visual grounding category includes Visual Haystack (image-only context) and MM-NIAH (interleaved image-text context). For Visual Haystack, the original dataset already provides a pre-filtered pool of up to 1,000 distractor images per instance. To circumvent the distractor collision issues discussed previously, we directly sample noise images from this curated pool to scale the inputs to our target length intervals. To evaluate the model's sensitivity to information placement, we explicitly insert the "needle" image at uniform depth intervals of 0\%, 20\%, 40\%, 60\%, 80\%, and 100\% within the sampled context. For MM-NIAH, the input length is calculated as the summation of text tokens and image tokens. We identify that MM-NIAH contains images with anomalously high token counts, which we compress to a maximum of 1,024 tokens. This normalization ensures that the length variations in our benchmark reflect genuine context scaling rather than artifactual token spikes.

For the video understanding tasks, LongVideoBench and Video-MME, sampling videos at a standard rate of 1 frame per second (fps) often leads to context lengths that far exceed 128K tokens. To ensure a balanced distribution of test cases across all four length intervals, we implement an adaptive sampling strategy based on video duration. For videos shorter than 60 seconds, we maintain a sampling rate of 1 fps to preserve temporal detail. Conversely, for videos exceeding 60 seconds, we reduce the sampling rate to 0.5 fps. This down-sampling effectively controls the context length within our target intervals without compromising the semantic continuity required for video comprehension.

\section{Human Evaluation}
\label{appendix:human_evaluation}

To rigorously validate the reliability and robustness of our automated LLM-as-a-judge evaluation paradigm, we conduct a comprehensive human evaluation to assess the alignment between human annotators and the automatic evaluator.

\paragraph{Experimental Setup} 
We select two representative models, Qwen2.5-VL-7B-Ins and Doubao-1.6, and sample instances across four diverse datasets encompassing our defined task categories: LongDocURL (Single-Source Visual QA), SlideVQA (Multi-Source Visual QA), Visual Haystack (Visual Grounding), and Video-MME (Video Understanding). Specifically, we randomly sample 20 instances from each dataset for both models, yielding a total of 160 cases ($2 \text{ models} \times 4 \text{ datasets} \times 20 \text{ instances}$). 

To ensure the high quality and professional standard of the annotations, three independent human annotators—all of whom are Ph.D. students specializing in multimodal research—are instructed to evaluate these cases following the exact same criteria as the automated judge. The citation metrics (Citation Precision and Citation Recall) are scored on a continuous scale, while the generation quality metric (Correctness) is assigned a discrete score. To derive the aggregated human ground truth, we compute the arithmetic mean of the three annotators' scores for the continuous citation metrics, and apply majority voting for the discrete correctness metric. 

To quantify the consistency among the three human annotators, we compute Krippendorff's $\alpha$. To measure the agreement between the aggregated human scores and the automatic evaluator, we utilize the Spearman rank correlation coefficient for the continuous citation metrics and Cohen's Kappa ($\kappa$) for the discrete correctness metric.

\paragraph{Experimental Analysis}
The evaluation results are summarized in Table~\ref{tab:human_agreement}. We first observe a remarkably high inter-annotator agreement across all metrics and models, with Krippendorff's $\alpha$ consistently exceeding 0.83. Notably, the $\alpha$ value for correctness on Doubao-1.6 reaches 0.9726, demonstrating the clarity and reproducibility of our evaluation criteria among humans. 

Furthermore, the aggregated human scores exhibit a strong positive correlation with the automatic evaluator's judgments. The Spearman correlation coefficients for Citation Precision and Recall are consistently above 0.90 for both models. Interestingly, we observe that the agreement between the aggregated human scores and the LLM judge is consistently higher than the agreement between any single human annotator and the judge. This indicates that our aggregation strategy—averaging continuous scores and majority-voting discrete scores—effectively mitigates individual annotation noise, yielding a more robust and reliable ground truth.

When analyzing the differences among the specific metrics, we find that the human-model agreement for Citation Recall is generally higher than that for Citation Precision (e.g., 0.9272 vs. 0.9095 on Qwen2.5-VL-7B). This discrepancy likely stems from the nature of the evaluation tasks: verifying whether a statement is supported by a set of sources (Recall) is a relatively straightforward cognitive task. In contrast, evaluating Citation Precision requires a rigorous "leave-one-out" assessment to determine the strict necessity of every single cited source, which inherently introduces more subjective judgment and complexity for human annotators. 

Overall, the near-perfect agreement scores (with Cohen's Kappa for correctness achieving up to 0.9393) substantiate that our LLM-based evaluator highly aligns with human perception and serves as a highly trustworthy proxy for assessing long-context multimodal faithfulness.

\begin{table}[h]
\centering
\caption{Human evaluation results. We report the inter-annotator agreement (Krippendorff's $\alpha$) among the three human evaluators, as well as the agreement between the aggregated human scores and the automated LLM judge (Spearman correlation for continuous metrics and Cohen's Kappa for discrete metrics).}
\label{tab:human_agreement}
\resizebox{\textwidth}{!}{
\begin{tabular}{lcccc}
\toprule
\multirow{2}{*}{\textbf{Metric}} & \multicolumn{2}{c}{\textbf{Inter-Annotator Agreement} (Krippendorff's $\alpha$ $\uparrow$)} & \multicolumn{2}{c}{\textbf{Human-Model Agreement} $\uparrow$} \\
\cmidrule(lr){2-3} \cmidrule(lr){4-5}
 & \textbf{Qwen2.5-VL-7B} & \textbf{Doubao-1.6} & \textbf{Qwen2.5-VL-7B} & \textbf{Doubao-1.6} \\
\midrule
Citation Precision & 0.8578 & 0.8361 & 0.9095 (Spearman) & 0.9005 (Spearman) \\
Citation Recall    & 0.9218 & 0.8977 & 0.9272 (Spearman) & 0.9196 (Spearman) \\
Correctness        & 0.9132 & 0.9726 & 0.9074 (Kappa)    & 0.9393 (Kappa) \\
\bottomrule
\end{tabular}
}
\end{table}

\section{Experimental Details of Attention Patterns}
\label{appendix:attention}
Since computing the attention scores requires storing the full attention matrix, which imposes substantial memory overhead, we restrict this analysis to Qwen-2.5-VL-3B and 7B, sampling 40 instances uniformly within the 0–16K context length range.
To quantify the model's reliance on relevant visual cues, we propose the Evidence Recall metric. We first aggregate attention scores between tokens of model responses and context images to identify the top-10 images with the highest values. Evidence Recall is then calculated as the number of ground-truth evidence images found within these top-10 images, divided by the total number of evidence images. This effectively measures the proportion of essential visual evidence successfully attended to by the model.

\begin{figure*}[t]
    \begin{AcademicBox}
You will receive a user's question about uploaded content (which could include text, images, video frames, etc.), a factual statement from an AI assistant's response based on that content, and part of the full content (since the full content is too long to display entirely). Your task is to carefully assess whether the provided content contains all the information of the statement. Please use the following grades to generate the rating:

- [[Relevant]] - All points of the statement are supported by the provided content.

- [[Irrelevant]] - The statement is unrelated to the provided content, or any points of the statement are inconsistent with the provided content.

Ensure that you do not use any information or knowledge outside of the provided content when evaluating.
    \end{AcademicBox}
    \vspace{-1em}
    \caption{Prompt for evaluating citation quality with GPT-5.2.}
    \label{fig:gpt4.1_evaluate_citation}
\end{figure*}

\begin{figure*}[t]
    \begin{AcademicBox}
You are asked to evaluate the quality of the AI assistant's answer to user questions as an impartial judge, and your evaluation should take into account factors including correctness (high priority), and comprehensiveness (whether the assistant's answer covers all points).

Read the AI assistant's answer and compare against the reference answer, and give an overall integer rating in 1, 2, 3 (1 = wrong or irrelevant, 2 = partially correct, 3 = correct and comprehensive) based on the above principles.
    \end{AcademicBox}
    \vspace{-1em}
    \caption{Prompt for evaluating correctness with GPT-5.2.}
    \label{fig:gpt4.1_evaluate_correctness}
\end{figure*}

\begin{figure*}[t]
    \begin{AcademicBox}
Instruction

Please answer the question with citations based only on the provided images. You must adhere to the following rules:

- Use an unbiased and journalistic tone.

- **Cite at least one and at most three images at the end of each sentence in your response.**

- Cite images using only their index number in brackets, for example [4][7], not [Image 4][Image 7].

- When citing several images, use separate brackets for each index number, like [a][b][c], instead of combining them like [a, b, c].

- If multiple images support a sentence, cite only a minimum sufficient subset of the images.
    \end{AcademicBox}
    \vspace{-1em}
    \caption{Instruction for image-only contexts.}
    \label{fig:image_only_ins}
\end{figure*}

\begin{figure*}[t]
    \begin{AcademicBox}
Instruction

Please answer the question with citations based only on the provided sources, which include both images and text passages. You must adhere to the following rules:

- Use an unbiased and journalistic tone.

- **Cite at least one and at most three sources at the end of each sentence in your response.**

- Cite sources using only their index number in brackets, for example [4][7], not [Image 4][Passage 7].

- When citing images or passages, use separate brackets for each index number, like [a][b][c], instead of combining them like [a, b, c].

- If multiple sources support a sentence, cite only a minimum sufficient subset of the sources.
    \end{AcademicBox}
    \vspace{-1em}
    \caption{Instruction for image-text contexts.}
    \label{fig:image_text_ins}
\end{figure*}

\begin{figure*}[t]
    \begin{AcademicBox}
Instruction

Please answer the question with citations based only on the provided video frames. You must adhere to the following rules:

- Use an unbiased and journalistic tone.

- **Cite at least one and at most three video frames at the end of each sentence in your response.**

- Cite frames using only their index number in brackets, for example [4][7], not [Frame 4][Frame 7].

- When citing several frames, use separate brackets for each index number, like [a][b][c], instead of combining them like [a, b, c].

- If multiple frames support a sentence, cite only a minimum sufficient subset of the frames.
    \end{AcademicBox}
    \vspace{-1em}
    \caption{Instruction for video-only contexts.}
    \label{fig:video_only_ins}
\end{figure*}

\section{Evaluation on \textsc{MMLongCite} across tasks}
\begin{figure*}[t]
    \centering
    \includegraphics[width=0.8\linewidth]{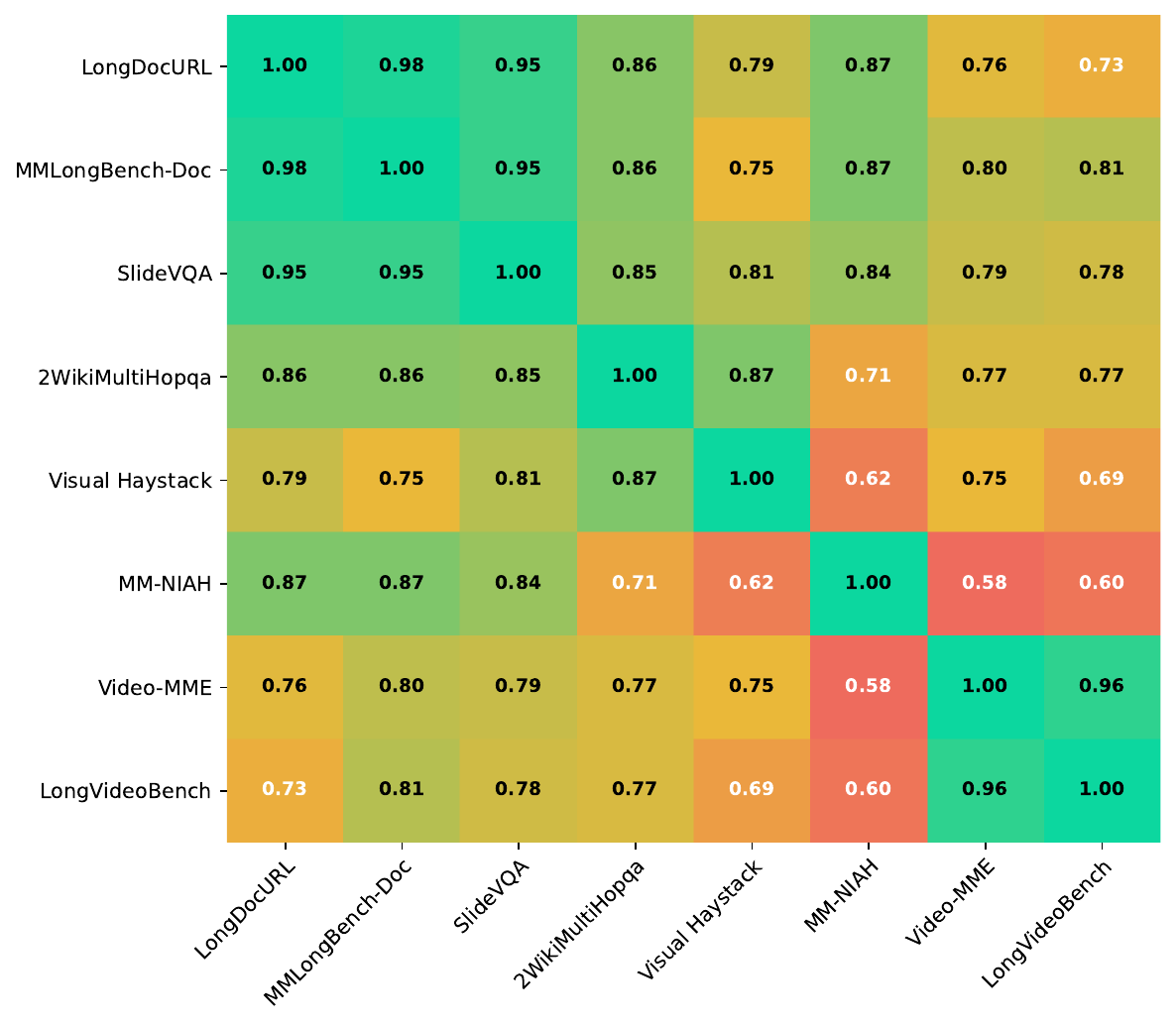}
    \caption{Correlation of Citation F1 across tasks in \textsc{MMLongCite}.}
    \label{fig:tasks_relation_citation}
\end{figure*}

\begin{figure*}[t]
    \centering
    \includegraphics[width=0.8\linewidth]{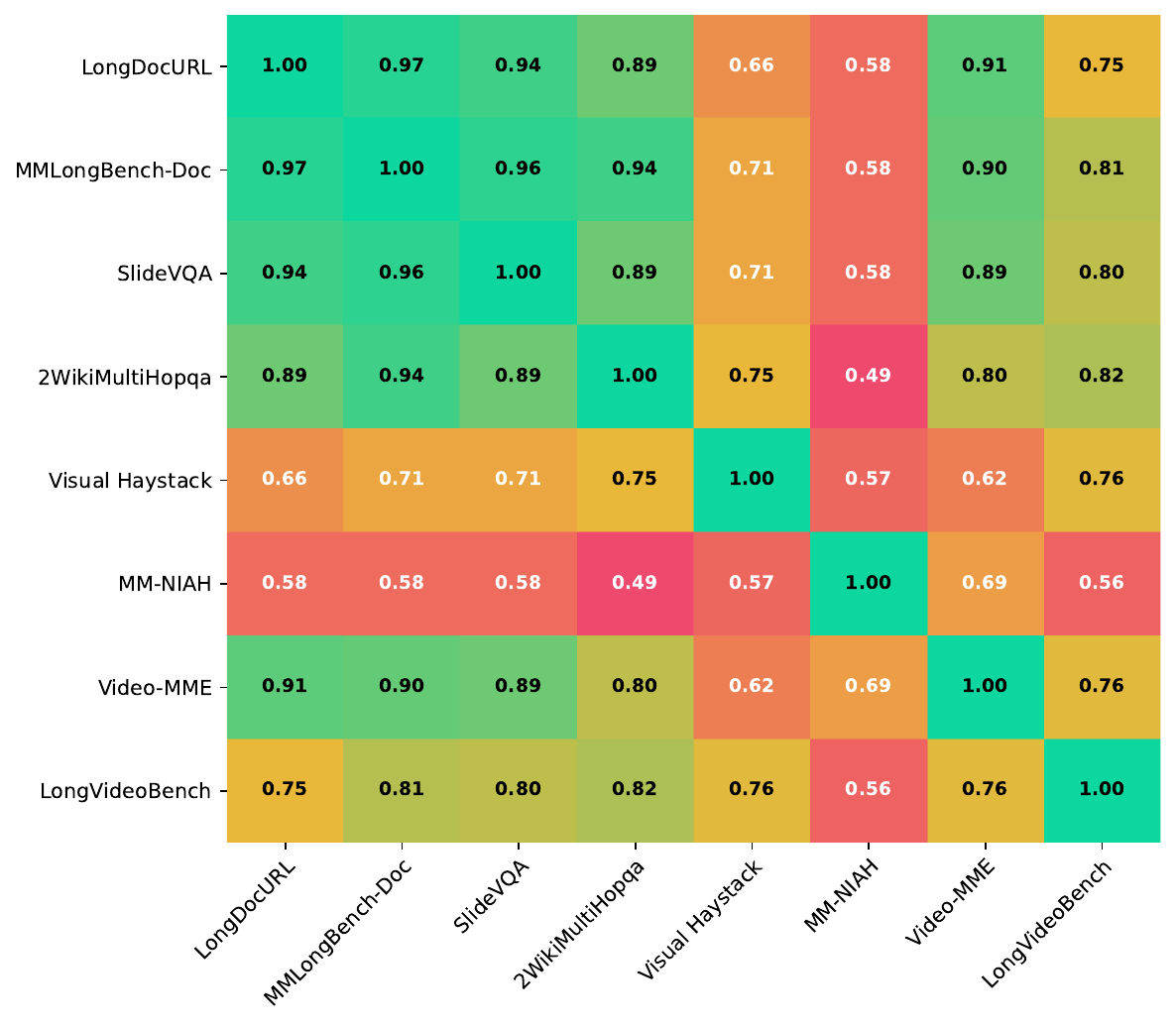}
    \caption{Correlation of Correctness across tasks in \textsc{MMLongCite}.}
    \label{fig:tasks_relation_correctness}
\end{figure*}

To systematically investigate the relationships and underlying capability dependencies among the evaluated tasks, we compute the Spearman correlation coefficients for both Citation F1 and Correctness metrics across all models. The resulting correlation matrices are visualized in Figure \ref{fig:tasks_relation_citation} and Figure \ref{fig:tasks_relation_correctness}.

\textbf{Intra-Task Consistency.} Homogeneous datasets within the same task category exhibit strong positive correlations, validating the reliability of our task taxonomy. For instance, within Single-Source Visual QA, LongDocURL and MMLongBench-Doc demonstrate exceptionally high correlations in Citation $F_1$ and Correctness. Similar consistency is observed in Multi-Source Visual QA and Video Understanding. This confirms that model performance remains highly stable when processing tasks with similar contextual structures and reasoning requirements.

\textbf{Inter-Task Divergence and Orthogonality.} Conversely, we observe pronounced divergence across heterogeneous task categories, indicating that distinct tasks evaluate orthogonal model capabilities. For example, the Citation $F_1$ correlation between Video Understanding (Video-MME) and Multi-Source QA (2WikiMultihopQA) drops notably, reflecting the mechanistic gap between temporal reasoning and cross-document information synthesis. This divergence is even more stark in the Correctness metric: the Visual Grounding task MM-NIAH exhibits remarkably low correlations with nearly all other tasks. Such variance underscores that excelling in one specific long-context skill (e.g., needle-in-a-haystack retrieval) does not trivially transfer to complex reasoning tasks, thereby justifying the necessity of encompassing diverse task categories within MMLongCite for a comprehensive evaluation.

\textbf{Decoupling of Faithfulness and Correctness.} A comparative analysis of the two metrics reveals a critical decoupling: models exhibit significantly higher cross-task consistency in attribution (Citation F1) than in standard accuracy (Correctness). Statistically, Citation F1 yields a higher average correlation (0.81) with a lower standard deviation (0.13) compared to Correctness (average: 0.76, standard deviation: 0.20). The heightened volatility in Correctness suggests that answering accuracy is highly susceptible to external factors and task-level variations. In contrast, the fundamental capability of grounding---the ability to trace and attribute information to the source context---remains a relatively invariant and stable model trait across different domains. This decoupling highlights the unique contribution of MMLongCite: relying exclusively on correctness metrics yields an inadequate assessment; incorporating explicit citation evaluation is imperative for a robust and trustworthy assessment of LCVLMs.

\begin{figure*}[t]
    \centering
    \includegraphics[width=\linewidth]{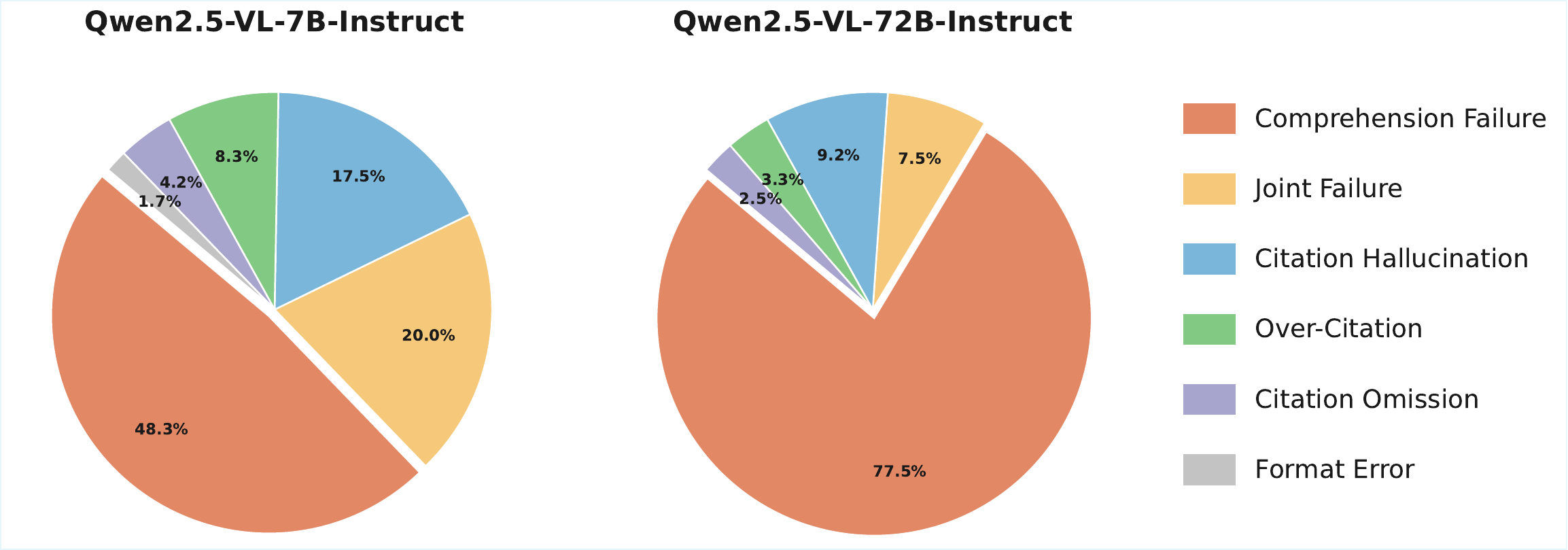}
    \caption{Error distribution in Qwen2.5-VL-7B-Instruct and Qwen2.5-VL-72B-Instruct.}
    \label{fig:error_analysis}
\end{figure*}

\section{Error Analysis}
\label{appendix:error_analysis}

To gain a deeper understanding of the bottlenecks in long-context multimodal citation, we conduct a fine-grained analysis of the model failure modes. We randomly sample 120 failure cases in total for each model, drawing 30 samples from each of the following four datasets: LongDocURL, MMLongBench-Doc, SlideVQA, and 2WikiMultiHopQA. Three independent PhD annotators are tasked with carefully reviewing the model predictions against the ground truth and supporting evidence, categorizing each failure into one of six predefined error types. 

\subsection{Error Taxonomy}
We defined six typical failure modes:
\begin{itemize}
    \item \textbf{Comprehension Failure:} The model successfully localizes the correct evidence (at least one cited index matches the ground truth) but fails to generate the correct answer.
    \item \textbf{Joint Failure:} The model completely fails at both tasks, providing an incorrect answer while citing entirely irrelevant images.
    \item \textbf{Citation Hallucination:} The model generates the correct answer (likely relying on its internal parametric knowledge or guessing) but fabricates citations or cites completely irrelevant images.
    \item \textbf{Over-Citation:} The model violates the citation density constraints by providing more than three citations within a single sentence, indicating a lack of precision in evidence selection.
    \item \textbf{Citation Omission:} The model generates a response but completely fails to provide any citation brackets throughout the entire output.
    \item \textbf{Format Error:} The model attempts to provide citations but violates the strict bracketed format requirement (e.g., using natural language descriptions instead of standard index brackets).
\end{itemize}

\subsection{Results and Analysis}
We analyze the error distributions of two representative models: Qwen2.5-VL-7B-Instruct and Qwen2.5-VL-72B-Instruct in Figure~\ref{fig:error_analysis}.

\textbf{Instruction Following and Formatting Capabilities.} Both models demonstrate strong adherence to formatting instructions. Citation Omission and Format Errors account for a negligible portion of the failures across both scales. Notably, the 72B model achieves a 0\% Format Error rate and reduces Over-Citation from 8.3\% (7B) to 3.3\%, indicating that scaling up significantly enhances the model's ability to strictly follow complex formatting and citation density constraints.

\textbf{The Shift from Localization Failures to Reasoning Bottlenecks.} The most striking observation is the drastic shift in the primary cause of failure between the two model scales. For the 7B model, localization-related errors (Joint Failure at 20.0\% and Citation Hallucination at 17.5\%) make up a substantial portion of the errors. This suggests that the smaller model frequently struggles to navigate the long multimodal context, either failing completely or relying on ungrounded parametric knowledge to answer the query.

In contrast, the 72B model significantly reduces Joint Failures (down to 7.5\%) and Citation Hallucinations (down to 9.2\%). However, this improvement in localization shifts the error distribution overwhelmingly toward \textbf{Comprehension Failure}, which surges from 48.3\% in the 7B model to 77.5\% in the 72B model. This indicates that while the larger model is highly proficient at retrieving the correct visual evidence from a massive context window, it still falls short in performing the complex, fine-grained cross-modal reasoning required to extract the exact answer from that localized evidence.

\section{Case Study}
\begin{figure*}[t]
    \centering
    \includegraphics[width=0.9\linewidth]{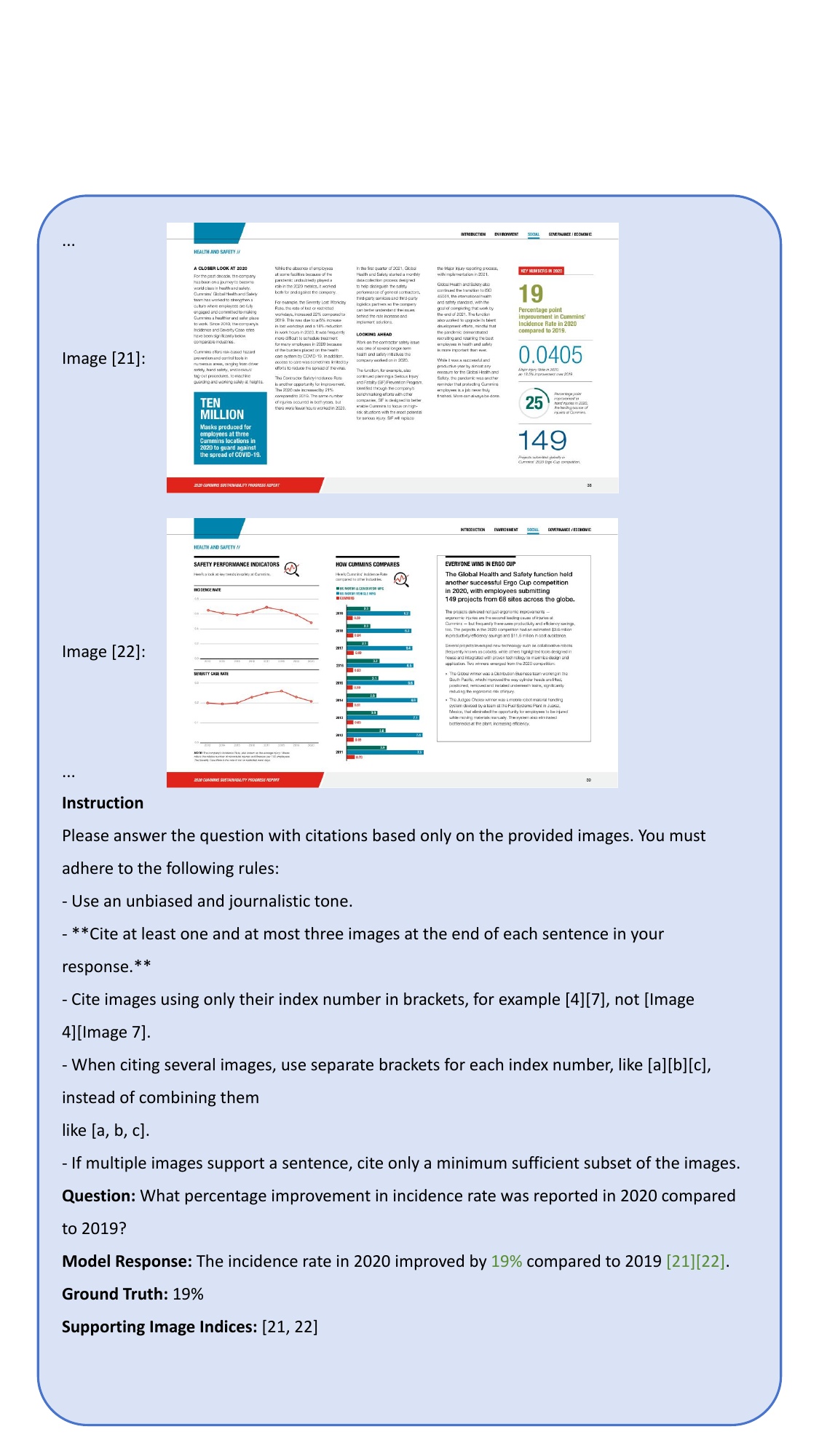}
    \caption{Case study of "Correct Case".}
    \label{fig:case_correct}
\end{figure*}

\begin{figure*}[t]
    \centering
    \includegraphics[width=0.9\linewidth]{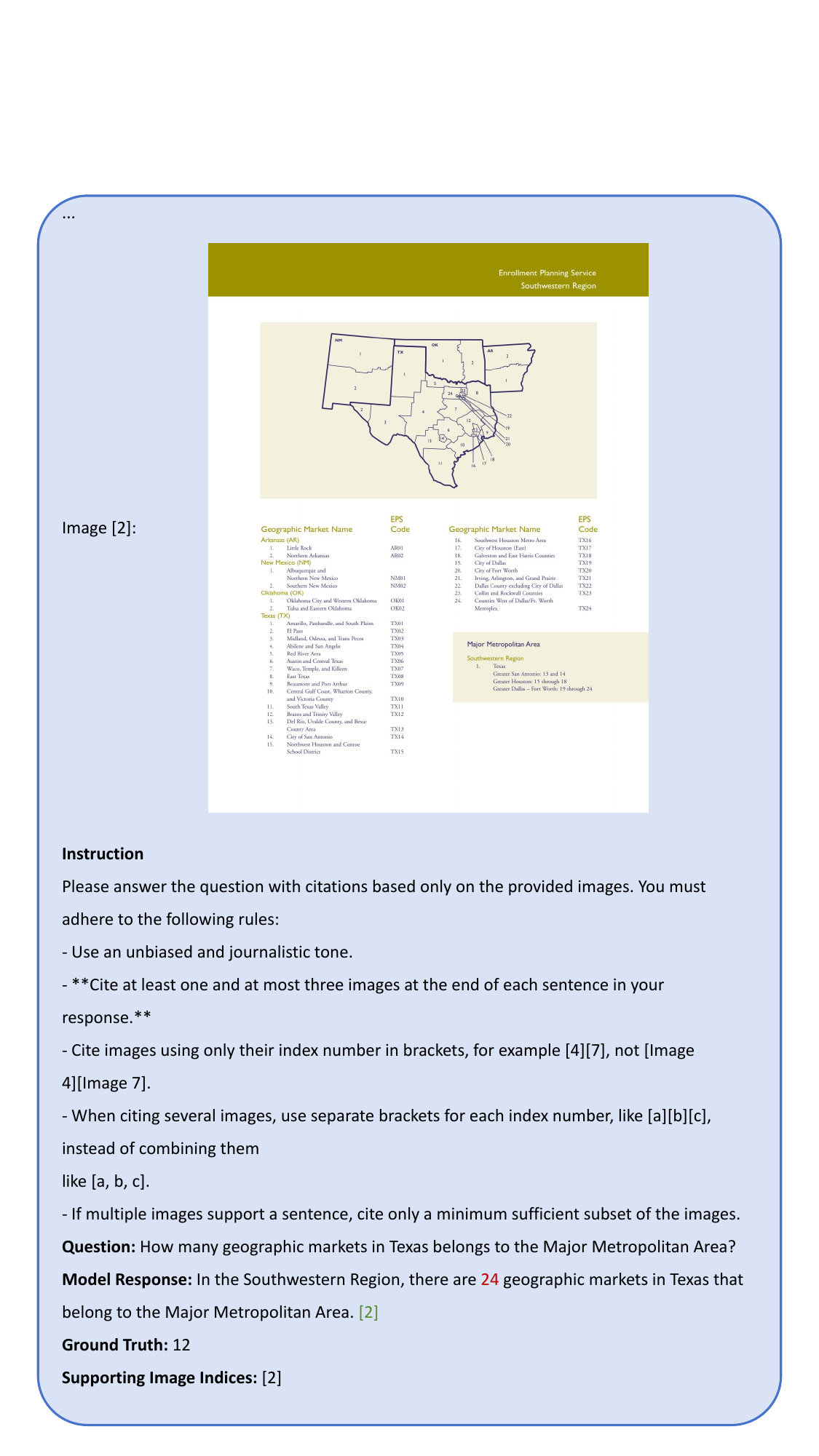}
    \caption{Case study of "Comprehension Failure".}
    \label{fig:case_comprehension}
\end{figure*}

\begin{figure*}[t]
    \centering
    \includegraphics[width=0.9\linewidth]{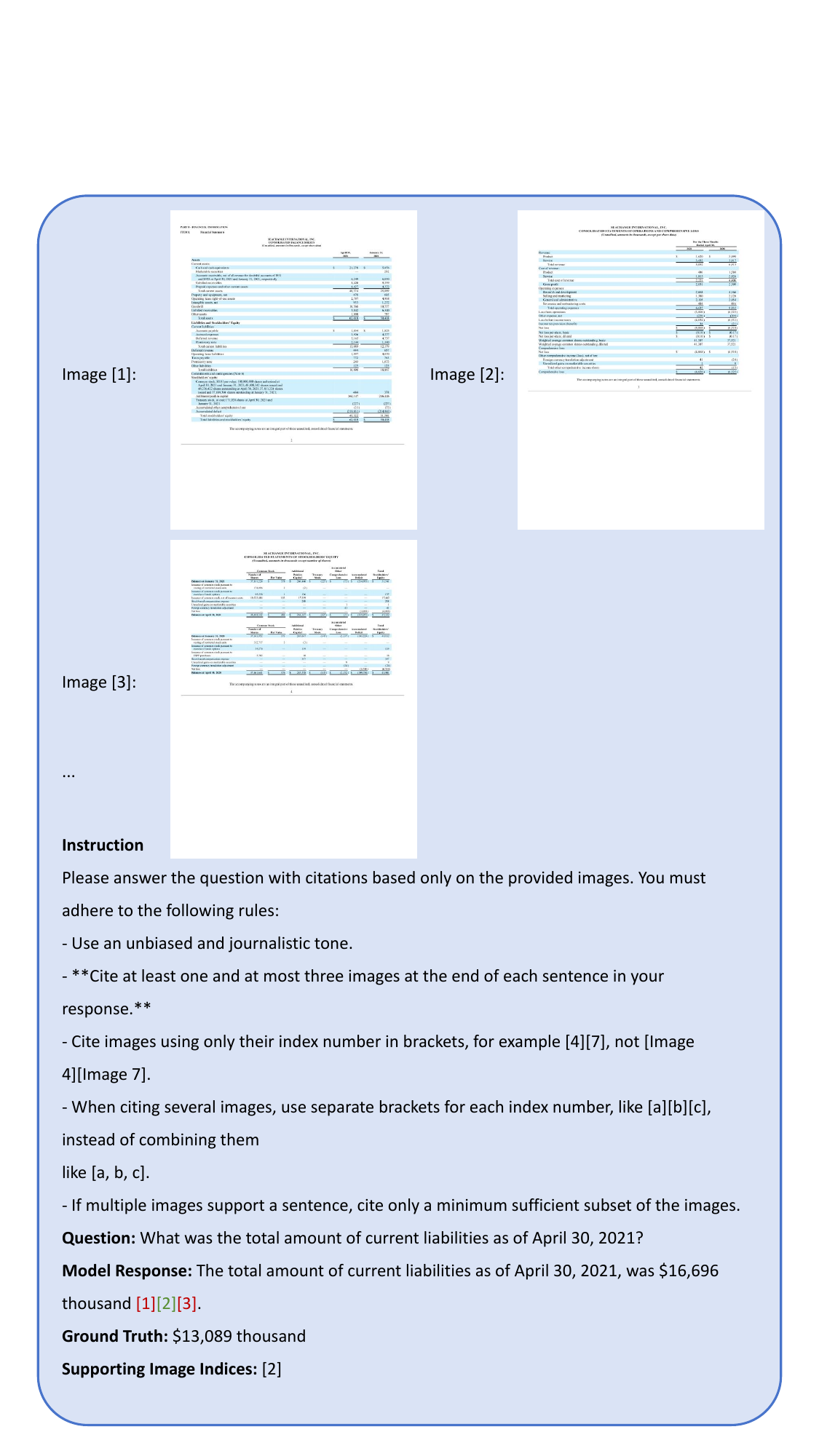}
    \caption{Case study of "Joint Failure".}
    \label{fig:case_joint_failure}
\end{figure*}

\begin{figure*}[t]
    \centering
    \includegraphics[width=0.9\linewidth]{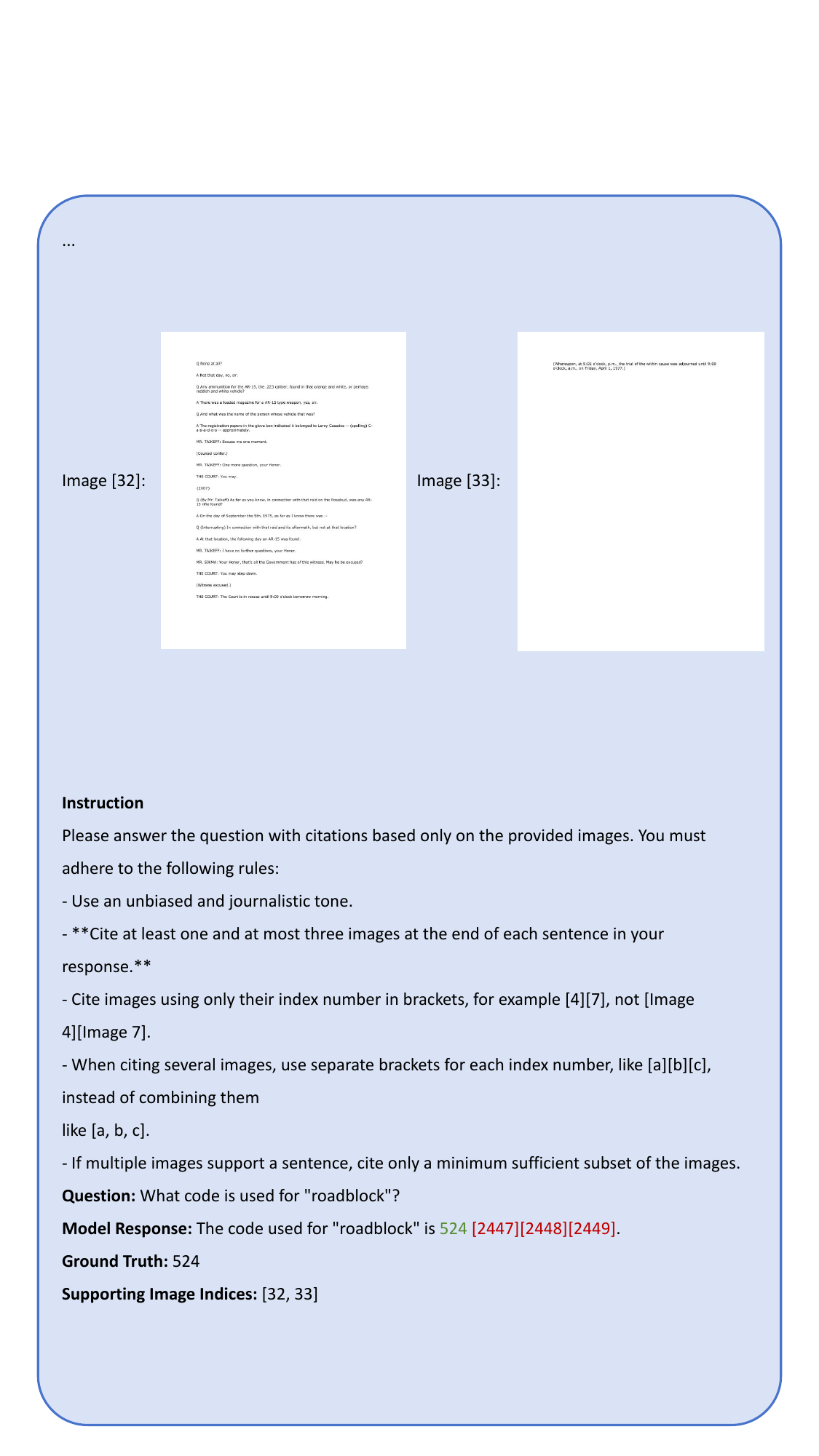}
    \caption{Case study of "Citation Hallucination".}
    \label{fig:case_citation_hallucination}
\end{figure*}

\begin{figure*}[t]
    \centering
    \includegraphics[width=0.9\linewidth]{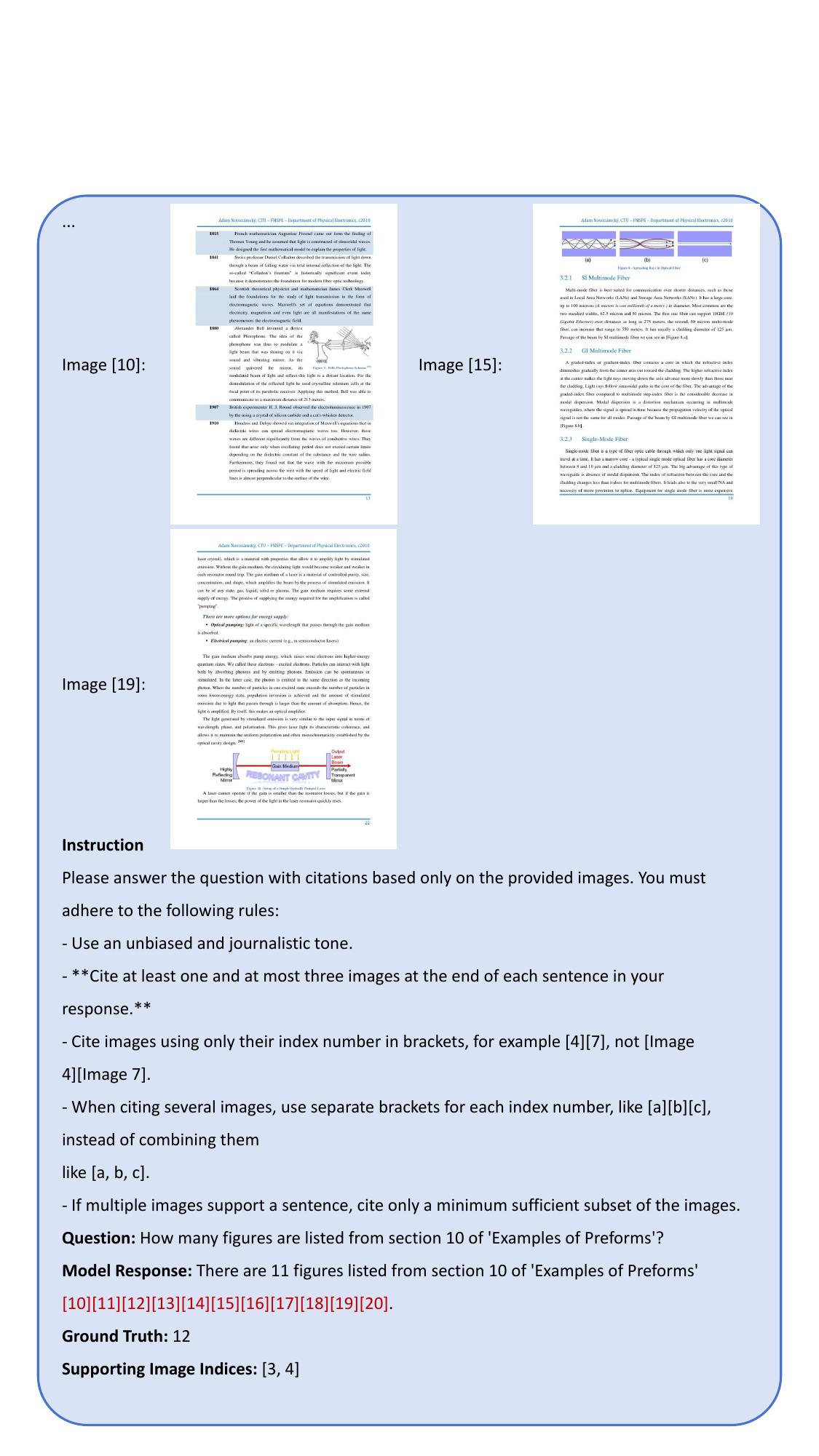}
    \caption{Case study of "Over Citation".}
    \label{fig:case_over_citation}
\end{figure*}

\begin{figure*}[t]
    \centering
    \includegraphics[width=0.9\linewidth]{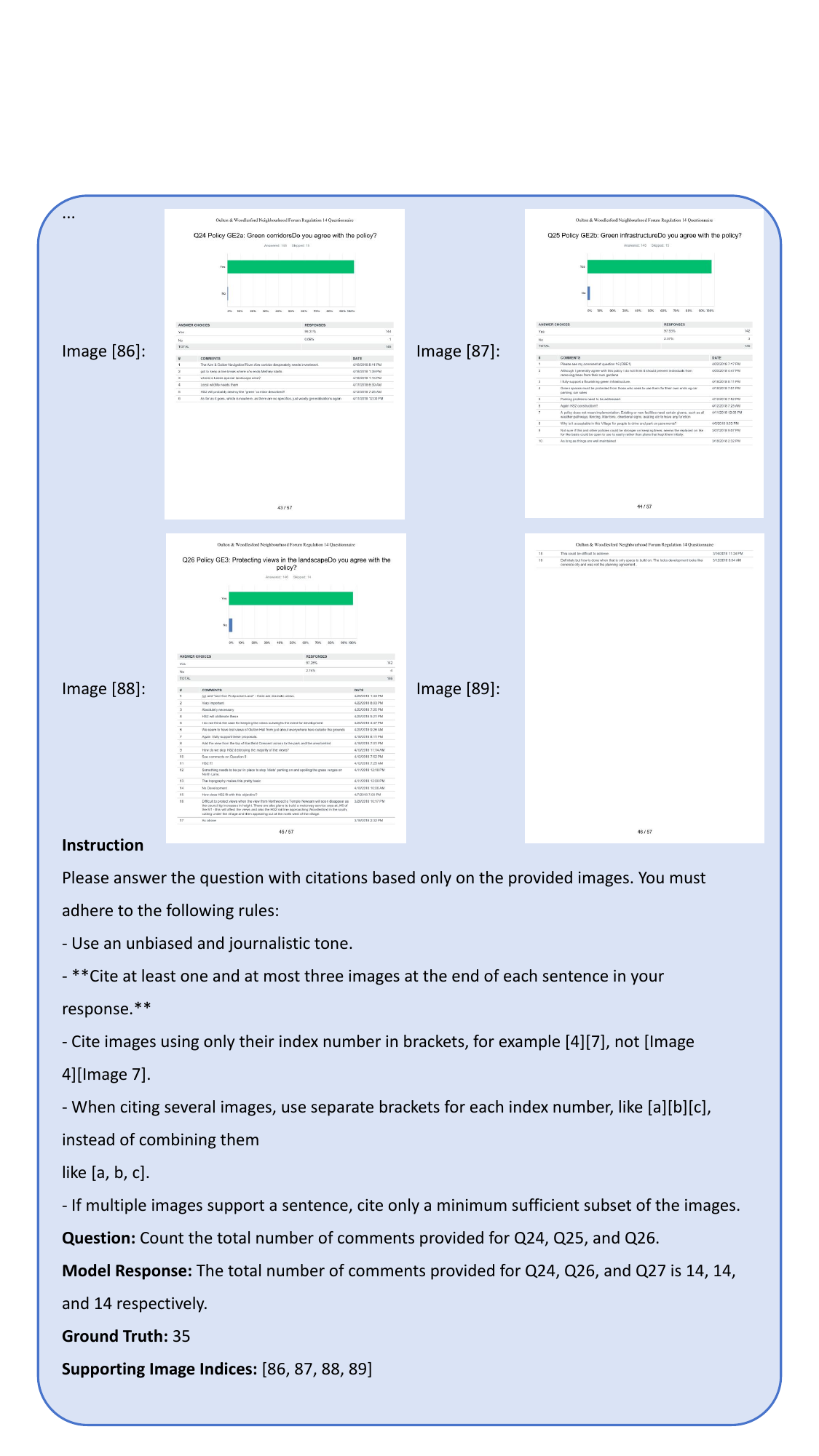}
    \caption{Case study of "Citation Omission".}
    \label{fig:case_citation_omission}
\end{figure*}

\begin{figure*}[t]
    \centering
    \includegraphics[width=0.9\linewidth]{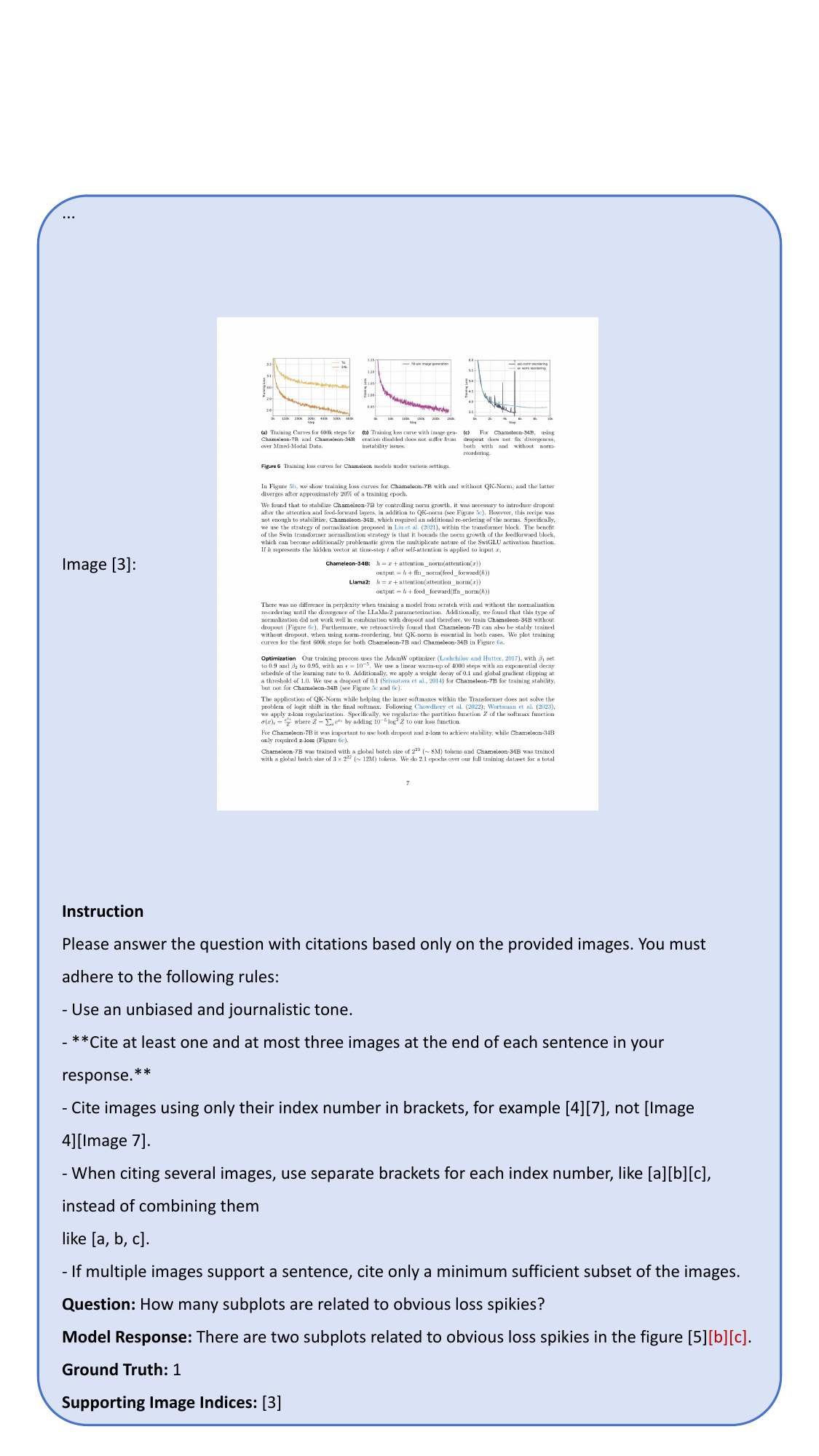}
    \caption{Case study of "Format Error".}
    \label{fig:case_format_error}
\end{figure*}

We list the instructions used for model inference from Figure~\ref{fig:image_only_ins} to Figure~\ref{fig:video_only_ins} and We present several qualitative cases from Figure~\ref{fig:case_correct} to Figure~\ref{fig:case_format_error}.

\section{Limitation}
\label{appendix:limitation}

\textsc{MMLongCite} uses automatic judges to evaluate citation
quality and answer correctness, which may introduce
judge-specific biases.
Although the second open-source judge yields consistent model rankings,
this does not eliminate all judge-specific biases.

The limited number of curated examples per condition constrains
distributional representativeness, particularly for video tasks with
only 40 examples per dataset--length cell.
Bootstrap confidence intervals quantify uncertainty within the current
samples but do not resolve this limitation.

The one-to-three citation constraint is an operational benchmark choice,
and absolute citation quality can vary with the cap.
Moreover, \textsc{MMLongCite-HR} measures both spatial reorganization
robustness and high-resolution visual grounding; naturally occurring
high-resolution inputs remain to be explored.
For \textsc{MMLongCite v2}, we plan to construct a new dataset
from scratch through dedicated data collection and annotation.
We aim to broaden source, domain, and modality coverage and
increase per-condition sample sizes.

\section{License}
\label{appendix:license}
All the data collected are based on previously open-sourced datasets, and all licenses are publicly available.

\section{Broader Impact}
\label{appendix:impact}
The long-context ability of LCVLMs has unlocked a large range of applications, including understanding lengthy multimodal documents, analyzing hour-long videos, and reasoning over complex, interleaved image-text data automatically. This ability, especially when augmented with faithful citation generation, can also help users to reliably summarize extensive visual archives, verify factual claims in dense reports, or trace answers back to specific visual evidence. Meanwhile, there are a large number of instruction-following scenarios grounded in complex vision-language contexts that require high reliability, such as evidence-based medical image analysis, legal document review, or transparent dialogue-based navigation for robots. Looking ahead, our MMLongCite will serve as a standard evaluation for the whole community to benchmark the faithfulness of new LCVLMs and to stimulate the development of models with more accurate cross-modal alignment, more reliable attention mechanisms for high-resolution visual grounding, and improved multimodal citation and reasoning capabilities.

\input{table/qwen_eva_res}

\section{Open-Source Judge Validation}
\label{appendix:open_source_judge}

To mitigate potential bias from relying on a single proprietary judge
and improve benchmark reproducibility, we use
Qwen3.6-27B\footnote{\url{https://huggingface.co/Qwen/Qwen3.6-27B}}
as a second, open-source judge to re-evaluate the benchmark outputs.
The results are reported in Table~\ref{tab:main_cite_qwen}.

Following the protocol in Appendix~\ref{appendix:human_evaluation},
we compare the judgments of GPT-5.2 and Qwen3.6-27B with the
aggregated human annotations on the same 160 cases.
Table~\ref{tab:judge_human_validation} shows that both judges
achieve high agreement with human annotations.
Although Qwen3.6-27B shows lower agreement on Doubao-1.6,
its agreement remains substantial, supporting its use as
a complementary open-source judge.

\begin{table}[t]
    \centering
    \small
    \caption{Agreement between each judge and the aggregated human
    annotations on the same 160 cases.
    CP and CR use Spearman correlation; Correctness uses Cohen's $\kappa$.}
    \label{tab:judge_human_validation}
    \begin{tabular}{llccc}
        \toprule
        Judge & Evaluated model & CP & CR & Correctness \\
        \midrule
        GPT-5.2      & Qwen2.5-VL-7B & 0.9095 & 0.9272 & 0.9074 \\
        Qwen3.6-27B  & Qwen2.5-VL-7B & 0.9322 & 0.9385 & 0.9124 \\
        GPT-5.2      & Doubao-1.6    & 0.9005 & 0.9196 & 0.9393 \\
        Qwen3.6-27B  & Doubao-1.6    & 0.8161 & 0.8489 & 0.7948 \\
        \bottomrule
    \end{tabular}
\end{table}

We further compare the model rankings and absolute scores
produced by the two judges.
Table~\ref{tab:judge_ranking_stability} reports Spearman and Kendall
rank correlations, together with the mean score shifts from
GPT-5.2 to Qwen3.6-27B.
The two judges produce nearly identical model rankings.
The average Citation F1 shift is only $+0.1$ percentage points,
whereas Qwen3.6-27B assigns Correctness scores $4.3$ points higher
on average, suggesting a difference in score calibration
rather than a change in comparative model conclusions.

\begin{table}[t]
    \centering
    \small
    \caption{Consistency between GPT-5.2 and Qwen3.6-27B.}
    \label{tab:judge_ranking_stability}
    \begin{tabular}{lccc}
        \toprule
        Metric & Spearman & Kendall & Qwen$-$GPT mean shift \\
        \midrule
        Citation F1 & 0.999 & 0.989 & $+0.1$ \\
        Correctness & 0.993 & 0.959 & $+4.3$ \\
        \bottomrule
    \end{tabular}
\end{table}

To examine potential near-family preference, we compare the
average score shifts between Qwen-family and non-Qwen models
when switching the evaluation judge from GPT-5.2 to Qwen3.6-27B.
Table~\ref{tab:judge_family_preference} shows that the shifts
are comparable between the two groups and do not consistently
favor Qwen-family models across metrics.
We do not observe a pronounced near-family preference
under this control.

\begin{table}[t]
    \centering
    \small
    \caption{Mean score changes when replacing GPT-5.2 with Qwen3.6-27B.}
    \label{tab:judge_family_preference}
    \begin{tabular}{lcc}
        \toprule
        Metric & Qwen-family & Non-Qwen  \\
        \midrule
        Citation F1 & $-0.05$ & $+0.16$ \\
        Correctness & $+4.82$ & $+3.90$ \\
        \bottomrule
    \end{tabular}
\end{table}

\section{Statistical Uncertainty Analysis}
\label{appendix:statistical_uncertainty}

We estimate 95\% confidence intervals using 10,000 bootstrap
resamples, sampling examples with replacement within each
context length interval of each dataset.
We report results for Gemini-3-Pro and Qwen2.5-VL-7B. Table~\ref{tab:task_confidence_intervals} reports the task-level
95\% confidence intervals.

\begin{table}[t]
    \centering
    \small
    \caption{Task-level scores with 95\% bootstrap confidence intervals.}
    \label{tab:task_confidence_intervals}
    \resizebox{\linewidth}{!}{
    \begin{tabular}{llcc}
        \toprule
        Model & Task & Citation F1 [95\% CI] & Correctness [95\% CI] \\
        \midrule
        \multirow{4}{*}{Gemini-3-Pro}
        & Single-Source Visual QA & 83.36 [81.19, 85.41] & 80.86 [77.97, 83.67] \\
        & Multi-Source Visual QA  & 84.02 [81.99, 85.90] & 85.85 [83.07, 88.46] \\
        & Visual Grounding       & 75.07 [72.74, 77.39] & 87.28 [85.08, 89.42] \\
        & Video Understanding    & 71.75 [68.64, 74.90] & 85.31 [81.25, 89.06] \\
        \midrule
        \multirow{4}{*}{Qwen2.5-VL-7B}
        & Single-Source Visual QA & 28.63 [25.69, 31.50] & 46.09 [42.42, 49.77] \\
        & Multi-Source Visual QA  & 33.10 [30.21, 35.96] & 52.59 [48.79, 56.36] \\
        & Visual Grounding       & 17.92 [15.85, 19.99] & 51.23 [47.77, 54.69] \\
        & Video Understanding    & 38.89 [34.51, 43.34] & 68.44 [63.28, 73.59] \\
        \bottomrule
    \end{tabular}}
\end{table}

\paragraph{Length-wise performance differences.}
We directly bootstrap the performance difference between the
64--128K and 0--16K bins; negative values indicate lower performance
at the longer context length.
Table~\ref{tab:length_confidence_intervals} shows greater length
sensitivity for Citation F1 than Correctness, particularly for
Qwen2.5-VL-7B on the non-video tasks.
Gemini-3-Pro exhibits smaller and more task-specific changes,
with its largest Citation F1 decline occurring on Visual Grounding.

For Video Understanding, the limited number of eligible source
examples within the required length ranges constrains the
per-dataset sample size.
We plan to increase the number of video examples in the next
version to obtain more precise and reliable performance
estimates across context lengths.
This limitation and the planned expansion are discussed in
Appendix~\ref{appendix:limitation}.

\begin{table}[t]
    \centering
    \small
    \caption{Performance differences between the 64--128K and 0--16K
    context bins, with 95\% bootstrap confidence intervals.}
    \label{tab:length_confidence_intervals}
    \resizebox{\linewidth}{!}{
    \begin{tabular}{llcc}
        \toprule
        Model & Task & $\Delta$ Citation F1 [95\% CI]
        & $\Delta$ Correctness [95\% CI] \\
        \midrule
        \multirow{4}{*}{Gemini-3-Pro}
        & Single-Source Visual QA
        & $-0.74\;[-6.56,\,5.05]$ & $+1.25\;[-7.19,\,9.69]$ \\
        & Multi-Source Visual QA
        & $-5.36\;[-9.33,\,-1.51]$ & $-1.74\;[-4.91,\,0.94]$ \\
        & Visual Grounding
        & $-20.53\;[-27.47,\,-13.58]$ & $-7.88\;[-14.94,\,-1.31]$ \\
        & Video Understanding
        & $-1.89\;[-10.30,\,6.64]$ & $+3.75\;[-7.50,\,15.00]$ \\
        \midrule
        \multirow{4}{*}{Qwen2.5-VL-7B}
        & Single-Source Visual QA
        & $-22.99\;[-31.15,\,-14.70]$ & $-5.63\;[-15.94,\,5.00]$ \\
        & Multi-Source Visual QA
        & $-35.34\;[-41.98,\,-28.70]$ & $-5.36\;[-12.10,\,1.21]$ \\
        & Visual Grounding
        & $-26.66\;[-35.33,\,-18.52]$ & $-21.12\;[-32.62,\,-8.69]$ \\
        & Video Understanding
        & $-0.81\;[-13.59,\,11.79]$ & $+0.62\;[-13.75,\,15.00]$ \\
        \bottomrule
    \end{tabular}}
\end{table}

\paragraph{Evidence-depth analysis.}
For the controlled single-needle subset of Visual Haystack,
we compare the average score at middle depths
(40\%, 60\%, and 80\% of the context) with that at edge depths
(0\% and 100\%).
We aggregate results across the four context length intervals
and use a cluster bootstrap to estimate 95\% confidence intervals.
A negative middle$-$edge difference indicates lower performance
for needles placed in the middle of the context.
As shown in Table~\ref{tab:depth_confidence_intervals},
Qwen2.5-VL-7B exhibits a larger decline in Citation F1 than
Gemini-3-Pro ($-6.42$ vs.\ $-2.52$).
The corresponding 95\% confidence interval for Qwen2.5-VL-7B
lies entirely below zero, supporting a middle-position decline
in citation quality.
The remaining confidence intervals in the table include zero,
providing less conclusive evidence of positional effects.

\begin{table}[t]
    \centering
    \small
    \caption{Middle$-$edge performance differences on the controlled
    single-needle subset of Visual Haystack, with 95\% cluster-bootstrap
    confidence intervals.}
    \label{tab:depth_confidence_intervals}
    \begin{tabular}{llc}
        \toprule
        Model & Metric & Middle$-$edge difference [95\% CI] \\
        \midrule
        Gemini-3-Pro  & Citation F1 & $-2.52\;[-7.65,\,3.81]$ \\
        Gemini-3-Pro  & Correctness & $-2.50\;[-9.17,\,3.33]$ \\
        Qwen2.5-VL-7B & Citation F1 & $-6.42\;[-15.75,\,-0.39]$ \\
        Qwen2.5-VL-7B & Correctness & $-1.67\;[-5.00,\,0.00]$ \\
        \bottomrule
    \end{tabular}
\end{table}

\section{Citation-Budget Sensitivity Analysis}
\label{appendix:citation_budget}

The default range of one to three citations per sentence is required
during both generation and evaluation as in ALCE~\citep{gao2023enabling}.
The upper bound encourages models to identify a compact sufficient
evidence set instead of citing many sources indiscriminately.
Nevertheless, three citations is an operational benchmark choice
rather than a universal rule for real-world writing.
We therefore evaluate sensitivity to this choice by varying the
maximum citations per sentence among 1, 3, 5, and 10.

We conduct a citation-budget ablation on all datasets using
Qwen2.5-VL-7B and Qwen3-VL-30B-A3B, varying the maximum citations
per sentence among 1, 3, 5, and 10 while retaining the lower bound of one.
Table~\ref{tab:citation_budget_sensitivity} reports the results.

The ordering between the two models remains unchanged under every
tested budget for all four metrics.
Relative to the default maximum of three citations, Correctness
changes by at most 0.82 percentage points.
Citation F1 shows modest model-specific variation at the more
restrictive or permissive budgets.
Thus, the comparison between these two models is robust across
the tested budgets, although absolute citation quality can vary
with the cap.

\begin{table}[t]
    \centering
    \small
    \caption{Citation-budget ablation across all eight datasets.}
    \label{tab:citation_budget_sensitivity}
    \begin{tabular}{lccccc}
        \toprule
        Model & Max citations & CP & CR & F1 & Correctness \\
        \midrule
        \multirow{4}{*}{Qwen2.5-VL-7B}
        & 1  & 33.69 & 28.86 & 30.37 & 54.31 \\
        & 3  & 30.75 & 32.24 & 29.64 & 54.59 \\
        & 5  & 30.49 & 31.32 & 29.00 & 54.62 \\
        & 10 & 28.54 & 30.41 & 27.30 & 55.25 \\
        \midrule
        \multirow{4}{*}{Qwen3-VL-30B-A3B}
        & 1  & 61.51 & 50.99 & 54.27 & 68.86 \\
        & 3  & 65.59 & 53.16 & 56.61 & 69.46 \\
        & 5  & 64.75 & 54.07 & 56.74 & 70.28 \\
        & 10 & 64.31 & 54.11 & 56.25 & 69.90 \\
        \bottomrule
    \end{tabular}
\end{table}

\section{Question-Only Evaluation and Contamination Analysis}
\label{appendix:question_only}

Models may answer questions from parametric knowledge without
using the supplied context.
To examine this possibility, we conduct a question-only evaluation
on four document-oriented datasets:
LongDocURL, MMLongBench-Doc, SlideVQA, and 2WikiMultiHopQA.
We provide each model with only the question, remove the entire
original context, and ask it to answer.
Table~\ref{tab:question_only} reports answer Correctness.

The question-only results suggest that some questions may be
answered from parametric knowledge without the supplied context.
Such knowledge is likely to affect Correctness more directly
than citation quality, since Citation Precision and Recall still
require mapping answer claims to sources in the supplied
benchmark context.


\begin{table}[t]
    \centering
    \small
    \caption{Question-only Correctness on four document-oriented
    datasets, with the entire original context removed.}
    \label{tab:question_only}
    \resizebox{\linewidth}{!}{
    \begin{tabular}{lcccc}
        \toprule
        Model & LongDocURL & \makecell{MMLongBench-\\Doc}
        & SlideVQA & 2WikiMultiHopQA \\
        \midrule
        Qwen2.5-VL-7B-Instruct
        & 16.56 & 5.00 & 15.71 & 9.38 \\
        Qwen3-VL-30B-A3B-Instruct
        & 19.84 & 8.44 & 18.21 & 20.00 \\
        \bottomrule
    \end{tabular}}
\end{table}

\section{Benchmark Construction Sensitivity}
\label{appendix:construction_sensitivity}

Benchmark construction choices may affect the capabilities measured
by the evaluation.
We examine sensitivity to the representation of 2WikiMultiHopQA,
document the MM-NIAH normalization and distractor-pool construction,
and audit the composite-image sizes used in \textsc{MMLongCite-HR}.

\paragraph{2WikiMultiHopQA rendering.}
We replace the rendered context images in 2WikiMultiHopQA with
their corresponding text passages and use those passages as
the context input.
We evaluate Qwen2.5-VL-7B and Qwen3-VL-30B-A3B under the same
generation and evaluation protocol.
Table~\ref{tab:rendering_sensitivity} compares Citation F1 and
Correctness under rendered-image and original-text inputs.

The two models respond differently to the representation change:
text input raises Citation F1 for Qwen2.5-VL-7B but lowers it
for Qwen3-VL-30B-A3B.
Within Multi-Source Visual QA, SlideVQA emphasizes understanding
native slide documents, whereas rendered 2WikiMultiHopQA
emphasizes OCR-based visual document understanding of
text-dominant content.

\begin{table}[t]
    \centering
    \small
    \caption{Sensitivity to context representation on 2WikiMultiHopQA.}
    \label{tab:rendering_sensitivity}
    \begin{tabular}{lcccc}
        \toprule
        \multirow{2}{*}{Model}
        & \multicolumn{2}{c}{Citation F1}
        & \multicolumn{2}{c}{Correctness} \\
        \cmidrule(lr){2-3}
        \cmidrule(lr){4-5}
        & Image & Text & Image & Text \\
        \midrule
        Qwen2.5-VL-7B    & 21.18 & 23.08 & 41.25 & 39.06 \\
        Qwen3-VL-30B-A3B & 59.36 & 53.75 & 61.72 & 62.97 \\
        \bottomrule
    \end{tabular}
\end{table}

\paragraph{MM-NIAH normalization.}
MM-NIAH includes non-text-rich images with resolutions
reaching up to $8{,}000 \times 6{,}000$ pixels.
These outliers consume disproportionately many visual tokens,
causing a single citation index to cover an overly large visual region
and making citation granularity too coarse.
We therefore preserve each image's aspect ratio while resizing
high-resolution outliers to a nominal budget of 1,024 visual tokens,
following a similar image-resizing strategy to
MMLongBench~\citep{wangmmlongbench}.

\paragraph{HR grid construction and image-size audit.}
We compare three input layouts using the same source images:
separate images, $2 \times 2$ grids containing four images,
and $4 \times 4$ grids containing sixteen images.
Each source image retains its original pixel dimensions during
grid construction.

To check whether composite inputs exceed the models'
image-processing limits, we compare the pixel area of each composite
model-input image with the configured limits of Qwen2.5-VL and Qwen3-VL.
Table~\ref{tab:hr_image_limits} reports the percentage exceeding
each limit.

\begin{table}[t]
    \centering
    \small
    \caption{Percentage of composite model-input images exceeding
    the configured image-size limits.}
    \label{tab:hr_image_limits}
    \begin{tabular}{lcc}
        \toprule
        Setting & Above Qwen2.5-VL limit & Above Qwen3-VL limit \\
        \midrule
        4-image grids  & 0.00\% & 0.00\% \\
        16-image grids & 4.11\% & 0.48\% \\
        \bottomrule
    \end{tabular}
\end{table}

Only a small fraction of sixteen-image grids exceed these limits.
However, this does not disentangle spatial reorganization
effects from high-resolution visual grounding.
Because \textsc{MMLongCite-HR} uses constructed image grids,
we interpret it as a dense compositional-input stress test.
A non-grid evaluation based on naturally occurring high-resolution
images, such as posters, remains future work.

%% file: table/model_statistics.tex
\begin{table}[t]
\centering
\small
\caption{Detailed information of models evaluated in \textsc{MMLongCite}.}
\resizebox{\linewidth}{!}{
\begin{tabular}{l c c c c c c}
\toprule
\bf Model & \bf Version & \bf Ctx. Size & \bf \#Param & \bf Architecture & \bf Open-source & \bf Think\\
\midrule
Gemini-3-Pro~\citep{gemini_3_pro_2025} & gemini-3-pro-preview & 1M & \faLock & \faLock & \xmark & \tmark \\
Gemini-3-Flash~\citep{gemini_3_flash} & gemini-3-flash-preview & 1M & \faLock & \faLock & \xmark & \tmark \\
Seed-2.0-Pro~\citep{bytedance_seed2_2026} & doubao-seed-2-0-pro-260215 & 256K & \faLock & \faLock & \xmark & \tmark \\
Seed-2.0-Lite~\citep{bytedance_seed2_2026} & doubao-seed-2-0-lite-260215 & 256K & \faLock & \faLock & \xmark & \tmark \\
Seed-2.0-Mini~\citep{bytedance_seed2_2026} & doubao-seed-2-0-mini-260215 & 256K & \faLock & \faLock & \xmark & \tmark \\
Seed-1.6~\citep{bytedance_seed1_6_2025} & doubao-seed-1-6-251015 & 256K & \faLock & \faLock & \xmark & \tmark \\
\midrule
Qwen2.5-VL-3B-Ins~\citep{bai2025qwen2} & Qwen2.5-VL-3B-Instruct & 128K & 3B & Dense & \cmark & \xmark \\
Qwen2.5-VL-7B-Ins~\citep{bai2025qwen2} & Qwen2.5-VL-7B-Instruct & 128K & 7B & Dense & \cmark & \xmark \\
Qwen2.5-VL-32B-Ins~\citep{bai2025qwen2} & Qwen2.5-VL-32B-Instruct & 128K & 32B & Dense & \cmark & \xmark \\
Qwen2.5-VL-72B-Ins~\citep{bai2025qwen2} & Qwen2.5-VL-72B-Instruct & 128K & 72B & Dense & \cmark & \xmark \\
Qwen3-VL-2B-Ins~\citep{Bai2025Qwen3VLTR} & Qwen3-VL-2B-Instruct & 256K & 2B & Dense & \cmark & \xmark \\
Qwen3-VL-4B-Ins~\citep{Bai2025Qwen3VLTR} & Qwen3-VL-4B-Instruct & 256K & 4B & Dense & \cmark & \xmark \\
Qwen3-VL-8B-Ins~\citep{Bai2025Qwen3VLTR} & Qwen3-VL-8B-Instruct & 256K & 8B & Dense & \cmark & \xmark \\
Qwen3-VL-30B-A3B-Ins~\citep{Bai2025Qwen3VLTR} & Qwen3-VL-30B-A3B-Instruct & 256K & 30B(3B) & MoE & \cmark & \xmark \\
Qwen3-VL-235B-A22B-Ins~\citep{Bai2025Qwen3VLTR} & Qwen3-VL-235B-A22B-Instruct & 256K & 235B(22B) & MoE & \cmark & \xmark \\
MiMo-VL-7B-SFT~\citep{coreteam2025mimovltechnicalreport} & MiMo-VL-7B-SFT-2508 & 128K & 7B & Dense & \cmark & \tmark \\
MiMo-VL-7B-RL~\citep{coreteam2025mimovltechnicalreport} & MiMo-VL-7B-RL-2508 & 128K & 7B & Dense & \cmark & \tmark \\
Kimi-VL-A3B-Ins~\citep{team2025kimi} & Kimi-VL-A3B-Instruct & 128K & 16B(3B) & MoE & \cmark & \xmark \\
Kimi-VL-A3B-Think~\citep{team2025kimi} & Kimi-VL-A3B-Thinking-2506 & 128K & 16B(3B) & MoE & \cmark & \cmark \\
InternVL3.5-4B~\citep{wang2025internvl3} & InternVL3\_5-4B-HF & 40K & 4B & Dense & \cmark & \tmark \\
Gemma-4-E4B~\citep{gemma_4} & gemma-4-E4B-it & 128K & 8B & Dense & \cmark & \tmark \\
Gemma-4-26B-A4B~\citep{gemma_4} & gemma-4-26B-A4B-it & 256K & 26B(4B) & MoE & \cmark & \tmark \\
GLM-4.6V-Flash~\citep{hong2025glm} & GLM-4.6V-Flash & 128K & 9B & Dense & \cmark & \tmark \\
GLM-4.6V~\citep{hong2025glm} & GLM-4.6V & 128K & 108B(12B) & MoE & \cmark & \tmark \\
Qwen3.5-2B~\citep{qwen3.5} & Qwen3.5-2B & 256K & 2B & Dense & \cmark & \tmark \\
Qwen3.5-4B~\citep{qwen3.5} & Qwen3.5-4B & 256K & 4B & Dense & \cmark & \tmark \\
Qwen3.5-9B~\citep{qwen3.5} & Qwen3.5-9B & 256K & 9B & Dense & \cmark & \tmark \\
Qwen3.5-122B-A10B~\citep{qwen3.5} & Qwen3.5-122B-A10B & 256K & 122B(10B) & MoE & \cmark & \tmark \\
\bottomrule
\end{tabular}}
\label{tab:model_statistic}
\end{table}

%% file: table/full_cite1.tex
\begin{table*}[t]
    \centering
    \caption{Performance of LCVLMs in Single-Source Visual QA and Multi-Source Visual QA of \textsc{MMLongCite}.}
    \resizebox{\textwidth}{!}{
    \begin{tabular}{l | cccc | cccc | cccc | cccc}
        \toprule
         \multirow{2}{*}{\bf Models} &  \multicolumn{4}{c|}{\bf LongDocURL } &  \multicolumn{4}{c|}{\bf MMLongBench-Doc} &  \multicolumn{4}{c|}{\bf SlideVQA } &  \multicolumn{4}{c}{\bf 2WikiMultiHopQA } \\
         \cmidrule{2-17}
         & \makecell[c]{$\mathbf{CP}$} & \makecell[c]{\bf $\mathbf{CR}$} & \makecell[c]{\bf $\mathbf{F_1}$}  & \makecell[c]{\bf $\mathbf{Cor}$}  &\makecell[c]{$\mathbf{CP}$} & \makecell[c]{\bf $\mathbf{CR}$} & \makecell[c]{\bf $\mathbf{F_1}$}  & \makecell[c]{\bf $\mathbf{Cor}$}  & \makecell[c]{$\mathbf{CP}$} & \makecell[c]{\bf $\mathbf{CR}$} & \makecell[c]{\bf $\mathbf{F_1}$}  & \makecell[c]{\bf $\mathbf{Cor}$}  & \makecell[c]{$\mathbf{CP}$} & \makecell[c]{\bf $\mathbf{CR}$} & \makecell[c]{\bf $\mathbf{F_1}$} & \makecell[c]{\bf $\mathbf{Cor}$}   \\
         \midrule
          \rowcolor{gray!20} \multicolumn{17}{c}{\textit{\textbf{Proprietary Models}}}  \\
Gemini-3-Pro & 88.07 & 87.35 & 87.07 & 85.78& 80.91 & 79.82 & 79.65 & 75.94& 91.05 & 92.25 & 91.35 & 88.57& 76.46 & 77.69 & 76.68 & 83.12 \\
Gemini-3-Flash & 82.03 & 87.04 & 83.81 & 87.50& 77.19 & 80.89 & 78.30 & 77.97& 89.97 & 94.10 & 91.48 & 90.18& 82.46 & 84.16 & 82.95 & 80.62 \\
Seed-2.0-Pro & 79.04 & 69.48 & 72.16 & 79.69& 73.47 & 67.27 & 68.98 & 67.66& 89.67 & 85.81 & 86.97 & 86.79& 73.72 & 62.87 & 66.16 & 70.62 \\
Seed-2.0-Lite & 56.31 & 54.27 & 53.90 & 77.44& 50.52 & 48.74 & 48.54 & 65.16& 67.95 & 69.30 & 67.74 & 89.11& 63.61 & 67.53 & 64.42 & 80.31 \\
Seed-2.0-Mini & 71.62 & 60.51 & 62.29 & 78.55& 65.84 & 52.46 & 55.90 & 61.25& 80.39 & 75.83 & 76.26 & 86.96& 58.94 & 54.45 & 53.57 & 66.72 \\
Seed-1.6 & 64.54 & 64.41 & 63.07 & 66.72& 64.25 & 62.72 & 62.64 & 60.62& 86.16 & 87.18 & 85.92 & 85.54& 67.42 & 68.35 & 65.86 & 70.47 \\
          \rowcolor{gray!20} \multicolumn{17}{c}{\textit{\textbf{Open-source Models}}}  \\
Qwen2.5-VL-3B-Ins & 4.01 & 4.79 & 3.97 & 38.44& 7.79 & 9.14 & 7.85 & 32.03& 13.06 & 16.88 & 13.74 & 53.93& 5.65 & 6.72 & 5.97 & 26.41 \\
Qwen2.5-VL-7B-Ins & 25.57 & 30.34 & 25.69 & 49.06& 32.03 & 34.08 & 31.58 & 43.12& 43.16 & 52.50 & 45.01 & 63.93& 23.31 & 20.70 & 21.18 & 41.25 \\
Qwen2.5-VL-32B-Ins & 41.38 & 31.07 & 34.02 & 58.91& 53.60 & 37.59 & 41.97 & 52.34& 64.20 & 49.89 & 54.05 & 74.64& 38.26 & 27.23 & 30.19 & 52.97 \\
Qwen2.5-VL-72B-Ins & 58.91 & 56.56 & 56.87 & 65.00& 56.69 & 56.09 & 55.75 & 54.84& 77.08 & 78.99 & 77.33 & 78.39& 42.89 & 42.89 & 42.20 & 42.34 \\
Qwen3-VL-2B-Ins & 20.17 & 20.57 & 19.23 & 41.09& 33.17 & 32.13 & 30.44 & 40.47& 34.97 & 40.91 & 35.63 & 56.61& 24.13 & 26.52 & 24.25 & 34.84 \\
Qwen3-VL-4B-Ins & 47.77 & 45.38 & 45.92 & 51.88& 59.35 & 52.45 & 54.60 & 52.34& 70.08 & 67.32 & 67.81 & 71.79& 45.39 & 41.49 & 42.51 & 45.16 \\
Qwen3-VL-8B-Ins & 60.40 & 52.97 & 55.32 & 65.62& 62.76 & 53.69 & 56.39 & 53.91& 75.36 & 66.20 & 68.79 & 73.93& 63.45 & 51.11 & 54.98 & 57.19 \\
Qwen3-VL-30B-A3B-Ins & 62.23 & 55.65 & 57.44 & 63.12& 66.78 & 59.74 & 61.50 & 57.81& 80.79 & 78.13 & 78.45 & 73.57& 67.50 & 55.94 & 59.36 & 61.72 \\
Qwen3-VL-235B-A22B-Ins & 69.85 & 66.75 & 67.19 & 72.97& 68.48 & 63.75 & 64.77 & 63.75& 85.67 & 85.42 & 84.50 & 84.11& 70.05 & 67.74 & 67.07 & 65.00 \\
MiMo-VL-7B-SFT & 48.62 & 43.71 & 44.50 & 55.47& 48.50 & 45.33 & 45.65 & 53.12& 74.65 & 73.39 & 72.87 & 75.89& 50.76 & 48.51 & 47.91 & 61.09 \\
MiMo-VL-7B-SFT(think) & 43.84 & 32.03 & 35.22 & 61.09& 50.76 & 39.05 & 42.19 & 57.34& 62.58 & 54.40 & 55.91 & 77.68& 64.32 & 53.14 & 56.09 & 67.34 \\
MiMo-VL-7B-RL & 55.71 & 44.62 & 47.96 & 59.22& 55.67 & 47.29 & 49.36 & 51.41& 63.12 & 61.73 & 61.13 & 72.32& 49.71 & 43.77 & 44.52 & 50.94 \\
MiMo-VL-7B-RL(think) & 54.66 & 32.95 & 38.93 & 64.06& 55.68 & 33.21 & 39.47 & 59.22& 68.68 & 52.00 & 56.46 & 79.11& 50.10 & 37.24 & 40.68 & 56.56 \\
Kimi-VL-A3B-Ins & 33.14 & 29.86 & 29.93 & 38.28& 24.53 & 22.27 & 22.27 & 28.91& 49.38 & 44.81 & 44.94 & 68.93& 39.35 & 30.56 & 33.20 & 40.78 \\
Kimi-VL-A3B-Think & 44.01 & 29.95 & 33.38 & 48.28& 45.28 & 27.25 & 31.96 & 42.50& 54.23 & 36.82 & 41.45 & 52.14& 47.35 & 33.05 & 36.22 & 41.56 \\
InternVL3\_5-4B & 5.48 & 4.70 & 4.71 & 13.75& 23.47 & 18.01 & 19.27 & 32.81& 12.64 & 12.58 & 12.11 & 22.86& 13.31 & 10.26 & 11.24 & 17.50 \\
Gemma-4-E4B & 40.81 & 44.94 & 42.23 & 42.97& 44.40 & 48.31 & 45.74 & 33.59& 66.67 & 71.85 & 68.43 & 56.61& 66.43 & 69.17 & 67.33 & 29.06 \\
Gemma-4-26B-A4B & 45.17 & 45.15 & 44.68 & 48.59& 52.38 & 53.41 & 52.28 & 42.81& 65.07 & 66.24 & 65.32 & 69.82& 64.35 & 66.96 & 65.33 & 32.97 \\
GLM-4.6V-Flash & 54.98 & 41.58 & 44.82 & 62.19& 57.83 & 45.28 & 48.60 & 56.88& 69.28 & 59.88 & 61.92 & 75.89& 49.40 & 44.63 & 44.20 & 53.44 \\
GLM-4.6V & 61.08 & 54.43 & 55.25 & 68.28& 56.44 & 51.69 & 52.73 & 60.78& 74.22 & 74.05 & 72.55 & 80.54& 58.72 & 56.06 & 54.83 & 63.75 \\
GLM-4.6V(think) & 66.76 & 61.79 & 62.64 & 74.38& 58.85 & 55.23 & 55.34 & 67.50& 73.62 & 78.33 & 74.07 & 80.71& 64.67 & 65.65 & 62.31 & 70.00 \\
Qwen3.5-2B & 37.80 & 27.13 & 30.22 & 51.72& 50.87 & 35.97 & 40.10 & 46.25& 59.97 & 48.36 & 51.50 & 70.89& 35.75 & 28.38 & 30.11 & 42.50 \\
Qwen3.5-4B & 70.55 & 60.45 & 63.69 & 68.75& 69.68 & 60.27 & 63.31 & 59.06& 81.54 & 74.64 & 76.83 & 77.68& 77.71 & 63.49 & 68.36 & 60.47 \\
Qwen3.5-9B & 70.02 & 54.00 & 58.36 & 73.44& 71.01 & 52.59 & 57.91 & 65.78& 84.23 & 73.57 & 76.53 & 86.25& 77.16 & 60.03 & 65.74 & 61.72 \\
Qwen3.5-122B-A10B & 80.05 & 71.48 & 73.72 & 77.19& 75.61 & 66.64 & 68.95 & 66.72& 92.54 & 87.50 & 88.53 & 88.39& 77.66 & 69.88 & 72.50 & 68.12 \\
         \bottomrule
    \end{tabular}}
    \label{tab:appendix_full_cite1}
\end{table*}

%% file: table/full_cite2.tex
\begin{table*}[t]
    \centering
    \caption{Performance of LCVLMs in Visual Grounding and Video Understanding of \textsc{MMLongCite}.}
    \resizebox{\textwidth}{!}{
    \begin{tabular}{l | cccc | cccc | cccc | cccc}
        \toprule
         \multirow{2}{*}{\bf Models} &  \multicolumn{4}{c|}{\bf Visual Haystack } &  \multicolumn{4}{c|}{\bf MM-NIAH} &  \multicolumn{4}{c|}{\bf Video-MME } &  \multicolumn{4}{c}{\bf LongVideoBench } \\
         \cmidrule{2-17}
         & \makecell[c]{$\mathbf{CP}$} & \makecell[c]{\bf $\mathbf{CR}$} & \makecell[c]{\bf $\mathbf{F_1}$}  & \makecell[c]{\bf $\mathbf{Cor}$}  &\makecell[c]{$\mathbf{CP}$} & \makecell[c]{\bf $\mathbf{CR}$} & \makecell[c]{\bf $\mathbf{F_1}$}  & \makecell[c]{\bf $\mathbf{Cor}$}  & \makecell[c]{$\mathbf{CP}$} & \makecell[c]{\bf $\mathbf{CR}$} & \makecell[c]{\bf $\mathbf{F_1}$}  & \makecell[c]{\bf $\mathbf{Cor}$}  & \makecell[c]{$\mathbf{CP}$} & \makecell[c]{\bf $\mathbf{CR}$} & \makecell[c]{\bf $\mathbf{F_1}$} & \makecell[c]{\bf $\mathbf{Cor}$}   \\
         \midrule
          \rowcolor{gray!20} \multicolumn{17}{c}{\textit{\textbf{Proprietary Models}}}  \\
         Gemini-3-Pro & 81.08 & 76.74 & 78.08 & 76.75& 79.88 & 68.34 & 72.05 & 97.81& 81.17 & 73.86 & 75.95 & 90.00& 74.78 & 64.92 & 67.56 & 80.62 \\
Gemini-3-Flash & 69.43 & 80.60 & 73.44 & 73.25& 87.65 & 91.98 & 89.20 & 95.78& 72.27 & 76.99 & 73.31 & 91.25& 72.52 & 71.49 & 70.17 & 78.12 \\
Seed-2.0-Pro & - & - & - & -& - & - & - & -& 72.72 & 47.50 & 55.40 & 89.17& 73.55 & 45.19 & 53.85 & 76.56 \\
Seed-2.0-Lite & - & - & - & -& - & - & - & -& 49.25 & 38.56 & 40.87 & 86.94& 52.23 & 42.10 & 44.58 & 72.50 \\
Seed-2.0-Mini & - & - & - & -& - & - & - & -& 68.90 & 39.67 & 48.33 & 85.35& 66.08 & 36.45 & 45.02 & 72.19 \\
Seed-1.6 & 62.08 & 58.08 & 59.48 & 62.88& 83.75 & 82.92 & 83.09 & 82.81& 69.34 & 70.32 & 67.57 & 83.75& 65.78 & 64.74 & 62.01 & 76.56 \\
          \rowcolor{gray!20} \multicolumn{17}{c}{\textit{\textbf{Open-source Models}}}  \\
Qwen2.5-VL-3B-Ins & 5.96 & 4.75 & 5.09 & 30.25& 3.54 & 2.55 & 2.76 & 55.78& 16.77 & 20.21 & 17.18 & 75.31& 28.65 & 30.91 & 27.82 & 57.19 \\
Qwen2.5-VL-7B-Ins & 31.48 & 27.96 & 28.08 & 52.62& 7.86 & 8.23 & 7.75 & 49.84& 41.86 & 44.90 & 40.87 & 71.56& 40.73 & 39.25 & 36.92 & 65.31 \\
Qwen2.5-VL-32B-Ins & 37.25 & 17.98 & 23.81 & 57.88& 50.42 & 38.43 & 42.16 & 67.03& 48.82 & 22.00 & 29.16 & 74.06& 54.11 & 31.71 & 37.50 & 65.31 \\
Qwen2.5-VL-72B-Ins & 63.01 & 48.62 & 53.27 & 48.00& 45.94 & 44.48 & 44.87 & 75.62& 68.38 & 56.79 & 59.40 & 81.56& 63.92 & 48.41 & 52.67 & 64.38 \\
Qwen3-VL-2B-Ins & 40.53 & 36.62 & 37.45 & 48.00& 18.59 & 18.84 & 17.67 & 73.12& 50.72 & 36.74 & 40.66 & 72.19& 55.90 & 41.15 & 45.62 & 57.19 \\
Qwen3-VL-4B-Ins & 54.66 & 33.87 & 40.80 & 45.75& 43.54 & 36.66 & 38.42 & 68.75& 65.16 & 54.55 & 56.65 & 75.31& 71.68 & 52.86 & 58.81 & 68.44 \\
Qwen3-VL-8B-Ins & 47.47 & 29.96 & 35.41 & 52.75& 36.74 & 30.47 & 32.09 & 62.34& 67.70 & 61.28 & 62.07 & 75.31& 73.64 & 60.13 & 63.36 & 62.50 \\
Qwen3-VL-30B-A3B-Ins & 66.61 & 43.69 & 51.05 & 60.25& 47.40 & 44.55 & 45.24 & 86.41& 64.78 & 42.76 & 48.40 & 79.06& 68.65 & 44.84 & 51.44 & 73.75 \\
Qwen3-VL-235B-A22B-Ins & 71.19 & 61.80 & 64.54 & 55.62& 76.17 & 61.62 & 66.13 & 77.03& 68.49 & 68.22 & 66.40 & 86.88& 71.38 & 57.43 & 61.28 & 74.38 \\
MiMo-VL-7B-SFT & 60.94 & 55.46 & 56.73 & 57.00& 25.75 & 24.40 & 24.21 & 52.81& 57.02 & 49.85 & 50.84 & 74.38& 53.40 & 48.43 & 48.29 & 66.88 \\
MiMo-VL-7B-SFT(think) & 54.17 & 46.76 & 48.69 & 55.50& 27.68 & 21.76 & 23.38 & 51.72& 40.22 & 21.15 & 26.19 & 75.94& 60.60 & 28.71 & 36.72 & 70.94 \\
MiMo-VL-7B-RL & 57.61 & 54.33 & 54.68 & 54.88& 32.89 & 27.73 & 28.61 & 55.47& 56.67 & 47.34 & 49.56 & 78.75& 55.46 & 42.28 & 45.62 & 65.94 \\
MiMo-VL-7B-RL(think) & 72.09 & 49.23 & 55.25 & 58.00& 41.48 & 30.18 & 33.34 & 53.91& 39.39 & 17.97 & 22.95 & 76.56& 59.26 & 26.61 & 34.25 & 70.31 \\
Kimi-VL-A3B-Ins & 33.36 & 29.62 & 30.75 & 59.62& 14.45 & 13.93 & 14.09 & 65.00& 26.73 & 20.75 & 22.08 & 65.00& 13.92 & 12.14 & 12.05 & 70.62 \\
Kimi-VL-A3B-Think & 38.34 & 29.62 & 32.41 & 50.12& 18.94 & 13.18 & 14.53 & 33.91& 40.80 & 17.71 & 23.26 & 67.81& 37.56 & 13.84 & 18.83 & 42.19 \\
InternVL3\_5-4B & - & - & - & -& - & - & - & -& 8.00 & 6.82 & 6.68 & 40.62& 22.13 & 21.12 & 20.85 & 46.88 \\
Gemma-4-E4B & 61.32 & 65.83 & 62.86 & 48.38& 27.90 & 28.24 & 27.70 & 60.31& 65.90 & 72.32 & 67.24 & 52.50& 63.15 & 65.17 & 62.75 & 38.44 \\
Gemma-4-26B-A4B & 57.51 & 62.93 & 59.19 & 51.50& 24.80 & 25.03 & 24.86 & 68.44& 67.18 & 68.85 & 65.44 & 73.75& 69.90 & 63.92 & 65.24 & 65.94 \\
GLM-4.6V-Flash & 52.53 & 35.68 & 41.25 & 58.12& 55.16 & 56.20 & 55.38 & 97.03& 46.98 & 25.80 & 31.28 & 79.38& 50.11 & 23.37 & 30.28 & 69.06 \\
GLM-4.6V & 62.45 & 52.44 & 55.76 & 60.62& 29.64 & 29.69 & 29.53 & 95.62& 50.23 & 53.65 & 50.10 & 81.25& 48.97 & 47.86 & 46.36 & 71.56 \\
GLM-4.6V(think) & 76.04 & 69.62 & 71.14 & 65.88& 31.98 & 31.98 & 31.95 & 94.38& 62.29 & 57.90 & 56.72 & 85.94& 60.16 & 47.16 & 50.11 & 70.62 \\
Qwen3.5-2B & 54.79 & 43.79 & 47.36 & 60.88& 18.22 & 11.47 & 13.05 & 72.66& 53.48 & 24.98 & 32.35 & 76.25& 55.60 & 33.41 & 39.53 & 67.50 \\
Qwen3.5-4B & 69.09 & 55.19 & 58.82 & 53.88& 57.85 & 50.64 & 52.92 & 62.19& 68.36 & 46.53 & 52.91 & 80.62& 73.41 & 49.86 & 56.52 & 73.75 \\
Qwen3.5-9B & 67.67 & 55.64 & 59.22 & 64.88& 63.33 & 49.66 & 53.38 & 72.34& 65.20 & 45.82 & 51.15 & 81.56& 57.42 & 51.71 & 51.93 & 73.12 \\
Qwen3.5-122B-A10B & 76.17 & 58.77 & 64.23 & 63.25& 80.67 & 65.95 & 70.48 & 87.66& 74.09 & 50.91 & 56.76 & 84.38& 72.46 & 50.84 & 56.99 & 75.94 \\

         \bottomrule
    \end{tabular}}
    \label{tab:appendix_full_cite2}
\end{table*}

%% file: table/qwen_eva_res.tex
\begin{table*}[t]
    \centering
    \caption{Performance of LCVLMs in \textsc{MMLongCite}
evaluated by Qwen3.6-27B.}
    \resizebox{\textwidth}{!}{
    \begin{tabular}{l | cccc | cccc | cccc | cccc}
        \toprule
         \multirow{2}{*}{\bf Models} &  \multicolumn{4}{c|}{\bf Single-Source Visual QA} &  \multicolumn{4}{c|}{\bf Multi-Source Visual QA} &  \multicolumn{4}{c|}{\bf Visual Grounding } &  \multicolumn{4}{c}{\bf Video Understanding } \\
         \cmidrule{2-17}
         & \makecell[c]{$\mathbf{CP}$} & \makecell[c]{\bf $\mathbf{CR}$} & \makecell[c]{\bf $\mathbf{F_1}$}  & \makecell[c]{\bf $\mathbf{Cor}$}  &\makecell[c]{$\mathbf{CP}$} & \makecell[c]{\bf $\mathbf{CR}$} & \makecell[c]{\bf $\mathbf{F_1}$}  & \makecell[c]{\bf $\mathbf{Cor}$}  & \makecell[c]{$\mathbf{CP}$} & \makecell[c]{\bf $\mathbf{CR}$} & \makecell[c]{\bf $\mathbf{F_1}$}  & \makecell[c]{\bf $\mathbf{Cor}$}  & \makecell[c]{$\mathbf{CP}$} & \makecell[c]{\bf $\mathbf{CR}$} & \makecell[c]{\bf $\mathbf{F_1}$} & \makecell[c]{\bf $\mathbf{Cor}$}   \\
         \midrule
          \rowcolor{gray!20} \multicolumn{17}{c}{\textit{\textbf{Proprietary Models}}}  \\
         Gemini-3-Pro(think) & 87.06 & 86.14 & 85.86 & 85.31& 83.49 & 85.01 & 83.90 & 91.07& 80.20 & 71.88 & 74.63 & 81.34& 78.71 & 69.56 & 72.19 & 86.72 \\
Gemini-3-Flash(think) & 82.95 & 87.20 & 84.34 & 86.80& 87.29 & 90.44 & 88.44 & 91.14& 82.65 & 87.81 & 84.54 & 78.44& 69.90 & 73.44 & 69.77 & 86.25 \\
Seed-2.0-Pro & 81.47 & 73.06 & 75.15 & 81.02& 84.97 & 76.60 & 79.29 & 84.90& - & - & - & -& 70.68 & 45.34 & 52.92 & 84.59 \\
Seed-2.0-Lite & 59.90 & 56.82 & 57.02 & 80.08& 68.70 & 70.37 & 68.63 & 89.62& - & - & - & -& 44.71 & 36.89 & 38.17 & 81.14 \\
Seed-2.0-Mini & 73.57 & 60.03 & 63.04 & 78.76& 73.11 & 66.26 & 67.10 & 81.52& - & - & - & -& 61.02 & 37.05 & 43.04 & 81.44 \\
Seed-1.6 & 71.02 & 70.21 & 69.39 & 73.12& 80.69 & 79.05 & 78.80 & 82.66& 65.27 & 63.48 & 64.05 & 72.17& 65.25 & 67.16 & 63.51 & 82.50 \\
\rowcolor{gray!20} \multicolumn{17}{c}{\textit{\textbf{Open-source Models}}}  \\
Qwen2.5-VL-3B-Ins & 6.29 & 8.61 & 6.78 & 44.92& 8.74 & 11.45 & 9.51 & 47.09& 2.71 & 2.44 & 2.44 & 42.58& 18.59 & 22.27 & 18.98 & 67.97 \\
Qwen2.5-VL-7B-Ins & 31.03 & 36.64 & 31.73 & 56.48& 33.78 & 37.19 & 33.64 & 61.12& 15.33 & 14.63 & 14.21 & 51.30& 37.80 & 39.80 & 36.36 & 70.78 \\
Qwen2.5-VL-32B-Ins & 53.54 & 37.56 & 42.04 & 66.09& 52.21 & 39.87 & 43.53 & 71.73& 34.66 & 22.60 & 26.29 & 62.61& 43.17 & 23.47 & 28.83 & 73.12 \\
Qwen2.5-VL-72B-Ins & 65.30 & 63.27 & 63.45 & 68.12& 62.86 & 62.25 & 62.06 & 66.96& 45.07 & 37.74 & 40.11 & 63.31& 62.80 & 50.17 & 53.22 & 75.16 \\
Qwen3-VL-2B-Ins & 29.49 & 29.21 & 27.44 & 51.80& 31.74 & 36.80 & 32.58 & 56.57& 26.27 & 24.95 & 24.49 & 61.11& 49.25 & 36.62 & 40.30 & 68.28 \\
Qwen3-VL-4B-Ins & 61.21 & 55.70 & 57.19 & 62.11& 60.30 & 56.22 & 57.36 & 66.92& 44.53 & 31.91 & 36.04 & 57.34& 66.69 & 51.84 & 55.53 & 74.38 \\
Qwen3-VL-8B-Ins & 70.88 & 60.00 & 63.32 & 69.14& 68.26 & 58.11 & 61.18 & 73.44& 40.90 & 29.18 & 32.74 & 56.67& 65.72 & 60.01 & 59.85 & 72.34 \\
Qwen3-VL-30B-A3B-Ins & 69.96 & 61.85 & 63.98 & 69.14& 75.83 & 67.82 & 70.05 & 75.86& 51.79 & 39.14 & 43.21 & 73.27& 64.38 & 42.40 & 47.80 & 79.53 \\
Qwen3-VL-235B-A22B-Ins & 74.13 & 70.37 & 70.91 & 75.16& 78.84 & 76.18 & 76.20 & 80.11& 66.73 & 56.53 & 59.45 & 63.23& 67.69 & 62.68 & 62.07 & 82.81 \\
MiMo-VL-7B-SFT & 57.62 & 53.34 & 53.67 & 61.80& 65.19 & 63.04 & 62.62 & 75.19& 36.88 & 34.34 & 34.54 & 52.19& 52.98 & 48.05 & 48.39 & 72.34 \\
MiMo-VL-7B-SFT(think) & 51.48 & 38.17 & 41.63 & 68.12& 65.44 & 55.05 & 57.48 & 77.81& 35.60 & 30.59 & 32.03 & 53.61& 47.41 & 24.45 & 30.22 & 75.00 \\
MiMo-VL-7B-RL & 61.83 & 50.88 & 53.96 & 63.91& 60.39 & 55.96 & 56.42 & 68.65& 38.12 & 35.26 & 35.53 & 53.20& 55.27 & 44.26 & 47.10 & 73.12 \\
MiMo-VL-7B-RL(think) & 59.04 & 34.21 & 40.92 & 71.41& 62.07 & 45.69 & 50.04 & 74.35& 49.20 & 35.74 & 39.05 & 55.92& 43.67 & 20.57 & 26.35 & 75.78 \\
Kimi-VL-A3B-Ins & 32.27 & 29.19 & 29.17 & 39.22& 45.76 & 38.45 & 40.12 & 61.79& 23.34 & 21.17 & 21.76 & 62.39& 18.86 & 15.89 & 16.11 & 70.00 \\
Kimi-VL-A3B-Think & 49.43 & 31.44 & 35.90 & 55.62& 48.29 & 33.82 & 37.31 & 53.12& 25.28 & 19.28 & 21.10 & 42.39& 37.33 & 14.38 & 19.34 & 60.31 \\
InternVL3\_5-4B & 15.27 & 12.57 & 13.20 & 29.30& 12.19 & 11.11 & 11.19 & 25.66& - & - & - & -& 13.38 & 13.11 & 12.57 & 44.53 \\
Gemma-4-E4B & 49.47 & 53.98 & 50.98 & 47.66& 70.27 & 73.92 & 71.50 & 49.53& 38.38 & 42.13 & 39.57 & 52.80& 60.83 & 67.76 & 61.92 & 49.38 \\
Gemma-4-26B-A4B & 53.03 & 53.67 & 52.68 & 55.23& 66.47 & 68.71 & 67.20 & 59.54& 34.71 & 38.08 & 35.68 & 59.39& 64.45 & 64.10 & 61.75 & 73.44 \\
GLM-4.6V-Flash & 61.26 & 46.48 & 50.19 & 68.59& 61.03 & 53.07 & 54.30 & 72.63& 48.22 & 38.11 & 41.33 & 77.94& 41.71 & 21.90 & 26.32 & 77.50 \\
GLM-4.6V & 63.19 & 57.59 & 58.27 & 72.42& 70.03 & 66.70 & 66.36 & 79.51& 46.10 & 40.41 & 42.25 & 78.09& 48.53 & 49.84 & 47.15 & 79.06 \\
GLM-4.6V(think) & 64.74 & 59.63 & 60.36 & 77.81& 71.75 & 72.20 & 70.10 & 80.66& 52.22 & 49.20 & 50.09 & 80.20& 59.95 & 54.29 & 53.60 & 81.41 \\
Qwen3.5-2B & 49.02 & 35.48 & 39.36 & 59.92& 48.24 & 38.70 & 41.31 & 65.32& 35.17 & 26.79 & 29.40 & 66.69& 52.41 & 28.09 & 34.57 & 73.91 \\
Qwen3.5-4B & 77.48 & 66.50 & 69.96 & 73.36& 80.18 & 70.42 & 73.74 & 73.53& 61.35 & 51.54 & 54.19 & 57.25& 68.73 & 48.00 & 53.54 & 79.53 \\
Qwen3.5-9B & 76.30 & 57.03 & 62.50 & 78.83& 82.72 & 67.97 & 72.73 & 80.20& 63.14 & 50.81 & 54.67 & 69.41& 60.82 & 49.82 & 51.84 & 80.47 \\
Qwen3.5-122B-A10B & 81.41 & 72.54 & 74.94 & 79.53& 86.87 & 80.35 & 82.22 & 84.48& 72.38 & 58.30 & 62.53 & 75.09& 70.46 & 49.79 & 54.98 & 82.81 \\

         \bottomrule
    \end{tabular}}
    \label{tab:main_cite_qwen}
\end{table*}